%% file: main.tex
\documentclass[10pt,twocolumn,letterpaper]{article}

\usepackage{iccv}
\usepackage{times}
\usepackage{epsfig}
\usepackage{graphicx}
\usepackage{amsmath}
\usepackage{amssymb}
\usepackage{bm}
\usepackage{caption}
\usepackage{subcaption}
\usepackage{booktabs}
\usepackage{color}
\usepackage{url}
\usepackage[numbers,sort,compress]{natbib}
\usepackage[accsupp]{axessibility}

\renewcommand{\vec}[1]{{\bm #1}}

\definecolor{yellow}{rgb}{0.85, 0.85, 0.0}
\definecolor{darkyellow}{rgb}{0.8, 0.8, 0.0}

\usepackage[pagebackref=true,breaklinks=true,letterpaper=true,colorlinks,bookmarks=false]{hyperref}

\iccvfinalcopy 

\def\iccvPaperID{7063} 

\ificcvfinal\fi

\begin{document}

\title{Synthesizing Diverse Human Motions in 3D Indoor Scenes}


\newcommand*{\affaddr}[1]{#1} 
\newcommand*{\affmark}[1][*]{\textsuperscript{#1}}
\newcommand*{\email}[1]{\small{\texttt{#1}}}

\author{
Kaifeng Zhao\affmark[1] \quad
Yan Zhang\affmark[1] \quad
Shaofei Wang\affmark[1] \quad
Thabo Beeler\affmark[2]  \quad
Siyu Tang\affmark[1]\\
\affaddr{\affmark[1]ETH Z\"urich} \quad \affaddr{\affmark[2]Google} \\
\email{\{kaifeng.zhao, yan.zhang, shaofei.wang, siyu.tang\}@inf.ethz.ch} \\ \email{thabo.beeler@gmail.com}
}

\ificcvfinal\thispagestyle{empty}\fi

\twocolumn[{%
\renewcommand\twocolumn[1][]{#1}%
\maketitle
\begin{center}
    \newcommand{\teaserwidth}{\textwidth}
    \centerline{\includegraphics[width=\linewidth]{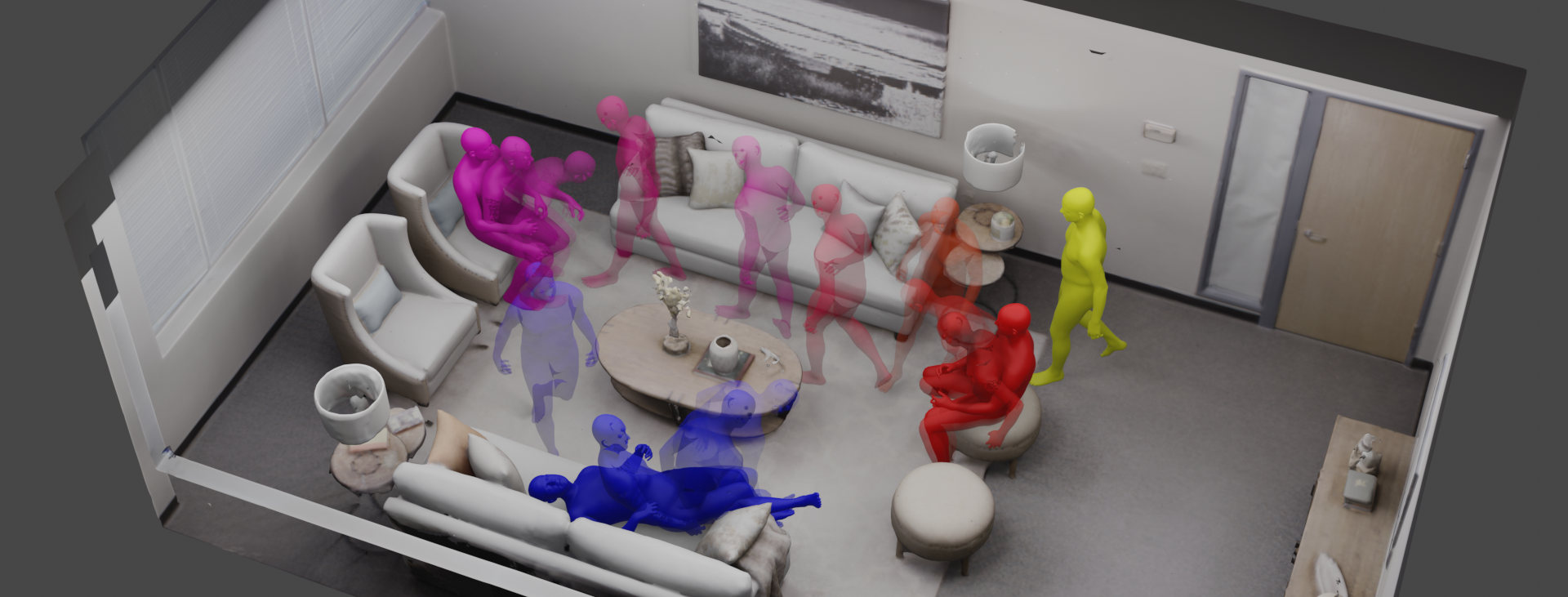}}
  \vspace{-0.1in}
  \captionof{figure}{
  In this work, we propose a method to generate a sequence of natural human-scene interaction events in real-world  complex scenes as illustrated here. The human first walks to sit on a stool (\textcolor{yellow}{yellow} to \textcolor{red}{red}), then walk to another chair to sit down (\textcolor{red}{red} to \textcolor{magenta}{magenta}), and finally walk to and lie on the sofa (\textcolor{magenta}{magenta} to \textcolor{blue}{blue}).}
\label{fig:teaser}
\end{center}%
}]

\maketitle



\input{sections/abstract.tex}
\input{sections/introduction.tex}
\input{sections/related.tex}

\input{sections/method/method.tex}

\input{sections/experiment/experiment.tex}

\input{sections/conclusion}

{\small
\bibliographystyle{ieee_fullname}
\bibliography{egbib}
}

\input{sections/supplementary}

\end{document}

%% file: sections/abstract.tex
\begin{abstract}
We present a novel method for populating 3D indoor scenes with virtual humans that can navigate in the environment and interact with objects in a realistic manner. 
Existing approaches rely on high-quality training sequences that contain captured human motions and the 3D scenes they interact with. However, such interaction data are costly, difficult to capture, and can hardly cover the full range of plausible human-scene interactions in complex indoor environments. 
To address these challenges, we propose a reinforcement learning-based approach that enables virtual humans to navigate in 3D scenes and interact with objects realistically and autonomously, driven by learned motion control policies.
The motion control policies employ latent motion action spaces, which correspond to realistic motion primitives and are learned from large-scale motion capture data using a powerful generative motion model. 
For navigation in a 3D environment, we propose a scene-aware policy with novel state and reward designs for collision avoidance. 
Combined with navigation mesh-based path-finding algorithms to generate intermediate waypoints, our approach enables the synthesis of diverse human motions navigating in 3D indoor scenes and avoiding obstacles.
To generate fine-grained human-object interactions, we carefully curate interaction goal guidance using a marker-based body representation and leverage features based on the signed distance field (SDF) to encode human-scene proximity relations. 
Our method can synthesize realistic and diverse human-object interactions (e.g.,~sitting on a chair and then getting up) even for out-of-distribution test scenarios with different object shapes, orientations, starting body positions, and poses. Experimental results demonstrate that our approach outperforms state-of-the-art human-scene interaction synthesis methods in terms of both motion naturalness and diversity.
Code, models, and demonstrative video results are available at: \href{https://zkf1997.github.io/DIMOS}{https://zkf1997.github.io/DIMOS}.
\end{abstract}

%% file: sections/introduction.tex
\section{Introduction}
Simulating how humans interact with environments plays an essential role in many applications, such as generating training data for machine learning algorithms, and simulating autonomous agents in AR/VR and computer games.  
Although this task is highly related to character animation in computer graphics, most existing character animation methods (e.g.~\cite{holden_learned_2020, buttner_motion_2015, buttner_machine_2019}) focus on improving the realism and controllability of character movements. With the traditional character animation workflows, one can produce high-quality animations but can hardly generate autonomous and spontaneous natural human motions interacting with the surroundings in diverse plausible ways as real humans. 
Previous learning-based interaction synthesis methods \cite{starke_neural_2019, hassan_stochastic_2021, zhang_couch_2022} require simultaneously capturing  human motion and scenes for supervision. However, capturing such training data is costly and challenging, resulting in a notably limited spectrum of human-scene interaction motions and difficulties in handling unseen interaction scenarios.
This restriction also results in inferior motion quality of synthesized virtual humans. 

To address this problem, we leverage reinforcement learning (RL)~\cite{sutton_introduction_1998} to solve our task. 
By formulating goals as rewards, perception as states, and latent variables of deep generative models as actions, we can synthesize continuous, stochastic, plausible, and spontaneous motions of virtual humans to inhabit the digital world. Although existing RL-based motion synthesis approaches (e.g.~\cite{ling_character_2020, zhang_wanderings_2022, peng_ase_2022}) can effectively generate natural motions to achieve goals, their generated virtual humans can only interact with simple scenes, rather than complex environments with functional furniture and diverse objects.
For example, GAMMA~\cite{zhang_wanderings_2022} employs generative motion primitives and a policy network that are generalizable across diverse human body shapes, but it can only synthesize waypoint-reaching locomotions. 
The trained digital humans are not aware of how to perform actions like sitting on a chair or lying on a sofa, and frequently inter-penetrate with the scene geometry.

To overcome these limitations, we propose a novel framework to learn both scene and interaction-aware motion control policies for synthesizing realistic and diverse human-scene interaction motions.
First, in order to improve the physical plausibility of the synthesized human motions, we design a new scene-aware policy to help virtual humans avoid collisions with scene objects. 
Specifically, we use a 2D occupancy-based local walkability map to incorporate scene information into the locomotion policy.
In addition, we add features derived from the signed distance from body markers to the object surface and the gradient direction of the signed distance  to encode the proximity between humans and objects for object interaction policies.
Second, in order to achieve controllable object interactions, we provide fine-grained guidances based on surface markers~\cite{zhang_we_2021} of a human body performing the target interactions. 
Specifically, we use COINS~\cite{zhao_compositional_2022} to generate human bodies interacting with scene objects given the interaction semantics, and then use the body markers as the interaction guidance for motion synthesis.
Combined with navigation mesh-based path-finding algorithms to generate intermediate waypoints in 3D scenes, virtual humans can autonomously reach target locations in complex environments and mimic target poses in a variety of plausible ways. 

We train the policy networks in synthetic scenes consisting of randomized objects to learn generalizable scene-aware locomotion and fine-grained object interactions. 
With this framework, we investigate how to synthesize diverse in-scene motions consisting of locomotion, sitting, and lying. We empirically evaluate the motion realism and expressiveness of our proposed method, and compare it with state-of-the-art methods. The results show that our approach consistently outperforms the baselines in terms of diversity, physical plausibility, and perceptual scores.

In summary, we aim to let virtual humans inhabit virtual environments, and present these contributions:
\begin{enumerate}
    \item We propose a reinforcement learning-based framework to generate realistic and diverse motions of virtual humans in complex indoor scenes.
    \item We propose to use body surface markers as detailed interaction goals for fine-grained human-object interaction synthesis and leverage COINS~\cite{zhao_compositional_2022} to generate articulated 3D human bodies based on interaction semantics to make virtual humans controllable via interaction semantics and fine-grained body poses.
    \item We design scene and interaction-aware policies to enable virtual humans to navigate in 3D scenes while avoiding collisions, to interact with scene objects, and to continuously perform sequences of activities in complex scenes.
\end{enumerate}

%% file: sections/related.tex
\section{Related Work}

\paragraph{Human motion synthesis.}
Generating high-quality human motion has been widely explored in computer vision and graphics.
Motion graph \cite{kovar_motion_2008} and motion matching \cite{buttner_machine_2019, buttner_motion_2015, zinno_ml_2019} generate motion by searching suitable clips in datasets and blending them automatically.
Starke \etal \cite{starke_neural_2019, starke_local_2020, starke_deepphase_2022} use phase-conditioned neural networks to synthesize character animations and interaction with objects.
Zhang \etal \cite{zhang_we_2021, zhang_wanderings_2022} model motion as sequences of surface markers on the parametric SMPL-X \cite{pavlakos_expressive_2019} body model, and train autoregressive networks on the large-scale mocap dataset AMASS~\cite{mahmood_amass_2019} in order to produce diverse motions of bodies with various shapes.
Peng \etal \cite{peng_ase_2022} use imitation learning with a skill discovery objective to learn a general motion skill space for physically simulated characters.
Tang \etal \cite{tang_real-time_2022} trains a motion manifold model of consecutive frames for real-time motion transition.
Recent works \cite{li_ganimator_2022, li_example-based_2023} propose generative methods to synthesize motions from single or a few example motion sequences.
Transformer-based models have been designed to predict or generate stochastic motions conditioned on action categories~\cite{petrovich_action-conditioned_2021}, texts~\cite{petrovich_temos_2022, tevet_motionclip_2022}, gaze~\cite{zheng_gimo_2022}, and others.
More recently, motion diffusion models ~\cite{tevet_human_2022, yuan_physdiff_2022, tseng_edge_2022, ao_gesturediffuclip_2023, alexanderson_listen_2023} achieve appealing performance on  motion synthesis conditioned on various control signals and demonstrate flexible motion editing.

\paragraph{Motion control and RL-based motion synthesis.}
Various motion control methods have been proposed to constrain body movements or guide the body to reach goals.
Sampling-based motion control methods \cite{liu_sampling-based_2010, liu_improving_2015} generate multiple samples at each step and select the samples that best match the targets.
Goal-conditional generation networks \cite{habibie_recurrent_2017, kania_trajevae_2021, holden_phase-functioned_2017} are applied for motion control.
However, such methods may produce invalid results when the train-test domain gap is large.
Optimization-based motion control methods \cite{holden_deep_2016, wang_combining_2019} leverage the learned generative motion model as regularization and optimize the motion latent variables to fit the decoded motion to the goals.
Motion diffusion models \cite{tevet_human_2022, yuan_physdiff_2022} implement control via classifier-free text guidance or gradually reprojecting the generated motion onto the physically plausible space at individual denoising steps.
Human motions can be formulated as a Markov decision process, and synthesized and controlled via RL.
Imitation learning methods \cite{peng_deepmimic_2018, bergamin_drecon_2019, peng_sfv_2018, merel_neural_2018} trains policy networks to control humanoids to imitate reference motion and complete given tasks.
Peng \etal~\cite{peng_ase_2022} propose a skill discovery objective apart from the imitation objective to learn a latent space for general motion skills and train substream task policies leveraging the general motion space. 
Follow-up works \cite{juravsky_padl_2022, tessler_calm_2023} combine the character controllers with language-based selection or finite state machine to compose more complex movements.
Ling \etal~\cite{ling_character_2020} and Zhang \etal~\cite{zhang_wanderings_2022} leverage the latent variable of the generative motion model as the action, and train policies to generate high-fidelity perpetual motions.

\paragraph{Human-scene interaction synthesis.}
Synthesizing natural human-scene interactions is an appealing direction in recent years.
Object affordance \cite{grabner_what_2011, savva_scenegrok_2014, hu_predictive_2020, gupta_3d_2011} provides cues about how humans can interact with scene objects.
Static bodies interacting with scene objects can be generated based on the scene semantics, geometries, and interaction semantics~\cite{li_putting_2019, hassan_populating_2021, zhang_place_2020, zhang_generating_2020, zhao_compositional_2022}.
Wang \etal \cite{wang_synthesizing_2021} propose to train generative models for human motion in scenes conditioned on scene pointclouds.
Wang \etal \cite{wang_towards_2022} propose a three-stage method to synthesize human motions in scenes, which first puts static bodies in scenes, then generates global paths, and afterwards synthesizes in-between motions.
Starke \cite{starke_neural_2019} propose an autoregressive neural state machine to synthesize character animations interacting with objects. 
Hassan \etal~\cite{hassan_stochastic_2021} and Zhang \etal~\cite{zhang_couch_2022} propose similar networks as in \cite{starke_neural_2019}, and incorporate stochastic motion generation and finer-grained motion control modules.
Peng \etal \cite{peng_ase_2022} trains policy to control physically-simulated characters to hit a box. Hassan \etal \cite{hassan_synthesizing_2023} extend the method to generate interactions like carrying a box or sitting on a chair. 
Zhang \etal \cite{zhang_learning_2023} learns physically simulated tennis players from broadcast videos.

\paragraph{Ours versus others.}
Our method is most similar to GAMMA~\cite{zhang_wanderings_2022} and SAMP~\cite{hassan_stochastic_2021}. 
GAMMA learns generalizable motion models and policies across human bodies of diverse identities and shapes, without goal-motion paired training data. Despite producing high-fidelity motions, its results are limited to locomotion in the scene, and frequently collide with scene objects.
SAMP learns conditional VAEs to produce sitting and lying actions in living environments, with object-motion paired data. Its generated motion has visible artifacts such as foot-ground skating.
Our method combines their merits and eliminates their individual disadvantages.
We extend the RL-based framework proposed in GAMMA by incorporating fine-grained motion controls (guided by interaction semantics) and scene interaction modules, so as to generate human-scene interactions in complex daily living environments.
Compared to SAMP, our produced motion is more diverse, more physically plausible, and can be guided by fine-grained body surface markers.
For systematic comparisons, please refer to Section~\ref{sec:experiment}.

%% file: sections/method/method.tex
\section{Method}

\input{sections/method/preliminary}

\input{sections/method/framework.tex}

\input{sections/method/locomotion.tex}
\input{sections/method/interaction.tex}


%% file: sections/method/preliminary.tex
\subsection{Preliminaries}
\paragraph{SMPL-X~\cite{pavlakos_expressive_2019} and body representation.}
We use SMPL-X to represent 3D human bodies in our work.
Given shape parameter $\vec{\beta} \in \mathbb{R}^{10}$ and body pose parameters $\vec{\theta} \in \mathbb{R}^{63}$, SMPL-X produces a posed body mesh with a fixed topology of 10475 vertices. To place the body in a scene, the root location $\vec{r}\in \mathbb{R}^3$ and the orientation $\vec{\phi} \in \mathfrak{so}(3)$ w.r.t. the scene coordinates are additionally needed.
Since facial expressions and hand gestures are not our focus, we leave their parameters as the default values. 
In addition, we follow~\cite{zhang_we_2021,zhang_wanderings_2022} to represent the body in motion by the $SSM\_{67}$ body surface marker placement. 
A motion sequence is then formulated as $\vec{X}=\{\vec{x}_1, ..., \vec{x}_N\}$, where $N$ is the length of motion and $\vec{x}_i \in \mathbb{R}^{67 \times 3}$ denotes the body marker 3D locations at frame $i$. 
The marker locations are relative to a canonical coordinate frame centered at the body pelvis in the first frame.

\paragraph{GAMMA~\cite{zhang_wanderings_2022}.}
GAMMA can synthesize stochastic, perpetual, and realistic goal-reaching actions in 3D scenes. It comprises generative motion primitive models, RL-based control, and tree-based search to implement gradient-free test-time optimization.
The motion primitive is formulated by a CVAE model to generate uncertain marker motions for 0.25 seconds into the future given a motion seed, followed by a MLP-based body regressor to yield SMPL-X parameters. Long-term random motion can be generated by running the motion primitive model recursively.
The RL-based control is implemented by learning a policy within a simulation area.
The actor-critic framework~\cite{sutton_introduction_1998} and the PPO algorithm~\cite{schulman_proximal_2017} are applied to update the policy network. An additional motion prior term is used to ensure the motion appears natural.
During testing, the generated motion primitives are stored in a tree where only the best K primitives at each layer are preserved in order to discard low-quality sampling results.

\paragraph{COINS\cite{zhao_compositional_2022}.}  COINS generates physically plausible static human-scene interactions with instance-level semantic control. 
Given the point cloud of an object and action labels, COINS can generate static bodies interacting with the given object based on the specified action, \eg, sitting, lying, or touching.  
COINS leverages transformer-based generative models trained on a human-scene interaction dataset~\cite{hassan_resolving_2019} to first generate a plausible body pelvis for interaction and then the posed body.  
The generated bodies are further optimized to improve the physical plausibility and to match the predicted action-dependent contact areas with objects.
Such generated static bodies capture the characteristics of the human-scene interaction process and can be used as fine-grained interaction guidance.

%% file: sections/method/framework.tex
\subsection{RL-based Framework to Inhabit the Virtual}
\begin{figure}[t]
\begin{center}
   \includegraphics[width=\linewidth]{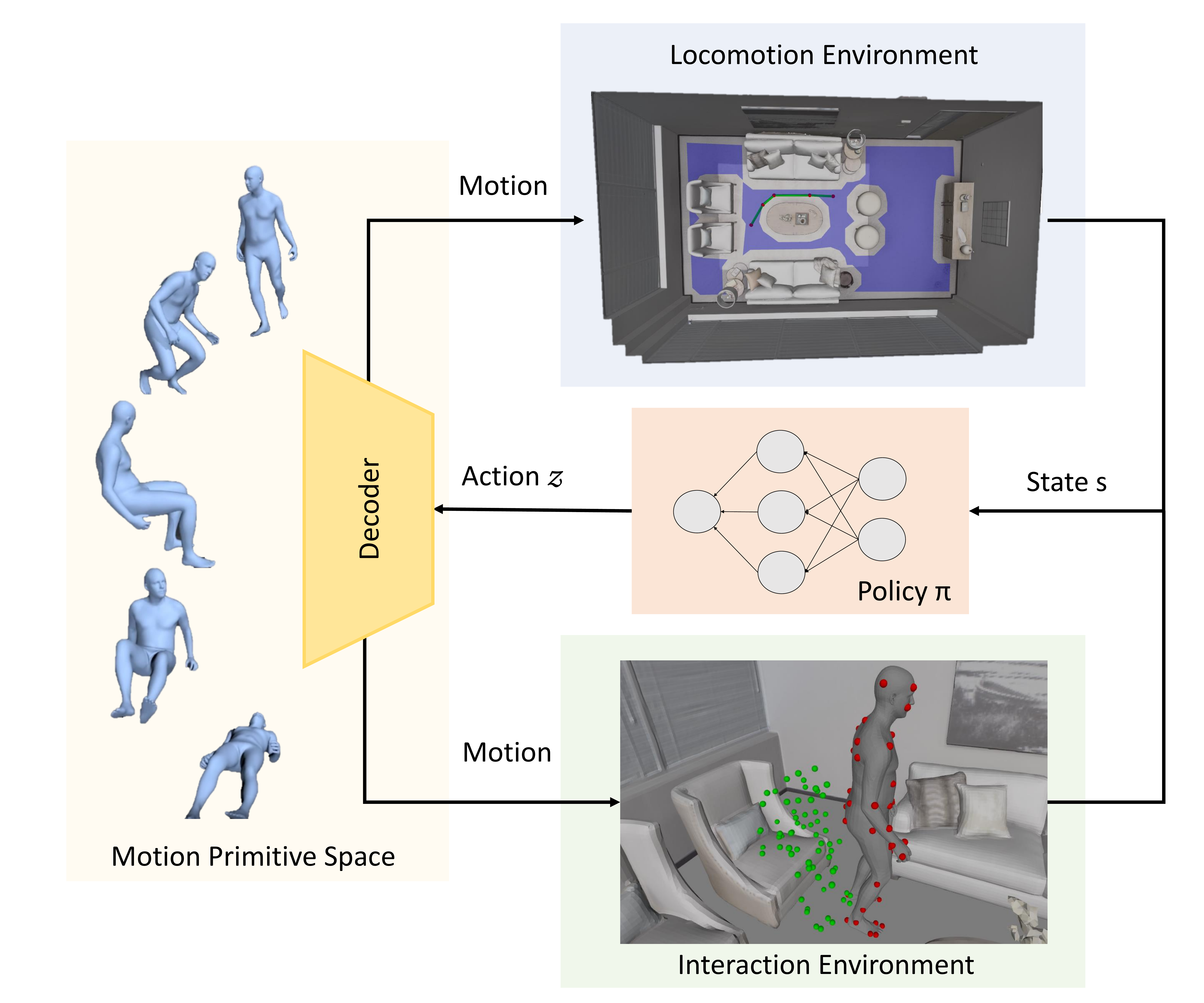}
\end{center}
\vspace{-5mm}
   \caption{
   Illustration of our proposed human-scene interaction synthesis framework, which consists of learned motion primitive (actions), alongside locomotion and interaction policies generating latent actions conditioned on scenes and interaction goals. 
   By integrating navigation mesh-based path-finding and static human-scene interaction generation methods, we can synthesize realistic motion sequences for virtual humans with fine-grained controls.
   }
\label{fig:overview}
\end{figure}

As illustrated in Fig.~\ref{fig:overview}, we propose a motion synthesis framework that enables virtual humans to navigate in complex indoor scenes and interact with various scene objects, \eg, sitting on a chair. 
Compared to the GAMMA framework~\cite{zhang_wanderings_2022}, 
our method incorporates scene information into the states to better handle complex human-scene interactions.
Also, we use body markers as goals to provide fine-grained guidance on how to drive the body for the target interactions.
With modularized path-finding methods and static person-scene interaction generation methods, our framework can synthesize realistic human motions in complex 3D environments. 
In our work, we use COINS~\cite{zhao_compositional_2022} to generate static person-scene interactions from interaction semantics given as `action-object' pairs. 
The walking path can be either generated by hand, or by automatic path-finding algorithms like A*~\cite{snook_simplified_2000, hart_formal_1968}.

We formulate our motion synthesis tasks with reinforcement learning.
At each time step, a virtual human perceives its state $\vec{s_t}$ in the environment and samples an action $\vec{a}_t$  from its policy model $\pi(\vec{a}_t|\vec{s}_t)$.
Based on its motion model, it advances its motion state, and obtains a new perception state $\vec{s_{t+1}}$. 
A reward $r_t=r(\vec{s}_t, \vec{a}_t, \vec{s}_{t+1})$ is calculated, tailored to different tasks. 

\paragraph{The motion model and the action.} 
We leverage the CVAE-based generative motion primitive~\cite{zhang_wanderings_2022} as our motion model, and use its latent variables as actions.
We train the model conditioned on 1 or 2 past frames using the combination of the SAMP~\cite{hassan_stochastic_2021} and AMASS~\cite{mahmood_amass_2019} motion capture datasets, to learn a latent motion primitive space covering motion skills for human-scene interactions.
Each latent variable $\vec{z}$ in the motion primitive space is regarded as an action and can be decoded to a short clip of motion.

\paragraph{The state.} 
The state is formulated by
\begin{equation}
\vec{s}_t = (\vec{X}_s, I, G),
\label{equ:abstrac_state}
\end{equation}
where $\vec{X}_s \in \mathbb{R}^{M \times 67 \times 3}$ is the body markers motion seed that represents a motion history of $M$ frames. 
$I$ and $G$ denote the person-scene interaction feature and the goal-reaching feature, respectively.
The interaction feature and goal-reaching feature vary among the locomotion and object interaction tasks.  We introduce the detailed formulation in Sec.~\ref{sec:locomotion} and \ref{sec:interaction}. 


\paragraph{The rewards.}
The rewards evaluate how well the virtual human performs locomotion and fine-grained object interaction tasks. 
We formulate rewards as
\begin{equation}
r = r_{goal} + r_{contact} + r_{pene},
\label{equ:abstract_reward}
\end{equation}
where $r_{goal}$, $r_{contact}$, and $r_{pene}$ represent the rewards for goal-reaching, foot-ground contact, and penetration avoidance, respectively.
Specifically, the contact reward $r_{contact}$ encourages foot-floor contact and discourages foot skating, and is defined as:
\begin{equation}
r_{contact} = e^{-(|\min x_z| - 0.05)_+} \cdot e^{-(\min\|x_{vel}\|_2 - 0.075)_+},
\label{equ:contact_reward}
\end{equation}
where $F$ is the set of foot markers, $\vec{x}_z$ is the height of the markers, $\vec{x}_{vel}$ is the velocity of the markers, and $(\cdot)_+$ denotes clipping negative values. There are tolerance thresholds of 0.05m for foot-floor distance and 0.075m/s for skating, following GAMMA~\cite{zhang_wanderings_2022}.
The other two reward terms are action-specific and introduced in Sec.~\ref{sec:locomotion} and \ref{sec:interaction}. 



\paragraph{Policy network and training.}
We use the actor-critic algorithm~\cite{sutton_introduction_1998} to learn the policy, where a policy network and a value network are trained jointly.
The policy network generates a diagonal Gaussian distribution representing the stochastic action distribution given a state, while the value network outputs the value estimation for each state. 

Like in GAMMA~\cite{zhang_wanderings_2022}, these two networks are jointly trained by minimizing:
\begin{equation}
\begin{split}
\mathcal{L} &= \mathcal{L}_{PPO} + \mathbb{E}[(r_t - V(\vec{s}_t))^2] \\
&+ \alpha \text{KL-div}(\pi(\vec{z}|\vec{s})||\mathcal{N}(0, I)),
\label{equ:loss_policy}
\end{split}
\end{equation}
where the first term is the PPO \cite{schulman_proximal_2017} loss, the second updates the value estimation of the critic networks, and the third Kullback–Leibler divergence term regularizes motion in the latent space~\cite{zhang_wanderings_2022}.

\paragraph{Tree sampling for test-time optimization.}
Given the stochastic nature of our Gaussian policies, sampling motions from the generated action distributions can yield motion primitive results of various qualities.
Therefore, we follow \cite{zhang_wanderings_2022} to use tree-based sampling during inference to discard motion primitives with inferior goal-reaching and scene interaction scores.
Specifically, we sample multiple latent actions at each time step and selectively keep the best K samples, utilizing the same rewards used to train the policies as the selection criteria.
This tree-sampling technique yields improved synthesis results of higher quality.

In the following, we elaborate on the design of states and rewards tailored to different actions, namely locomotion and fine-grained object interaction.
By combining the learned policies, long-term and coherent motions can be composed by rolling out the initial body and switching between the locomotion and object interaction stages.

%% file: sections/method/locomotion.tex
\subsection{Scene-Aware Locomotion Synthesis}
\label{sec:locomotion}
\begin{figure}[t]
\begin{center}
   \includegraphics[width=\linewidth]{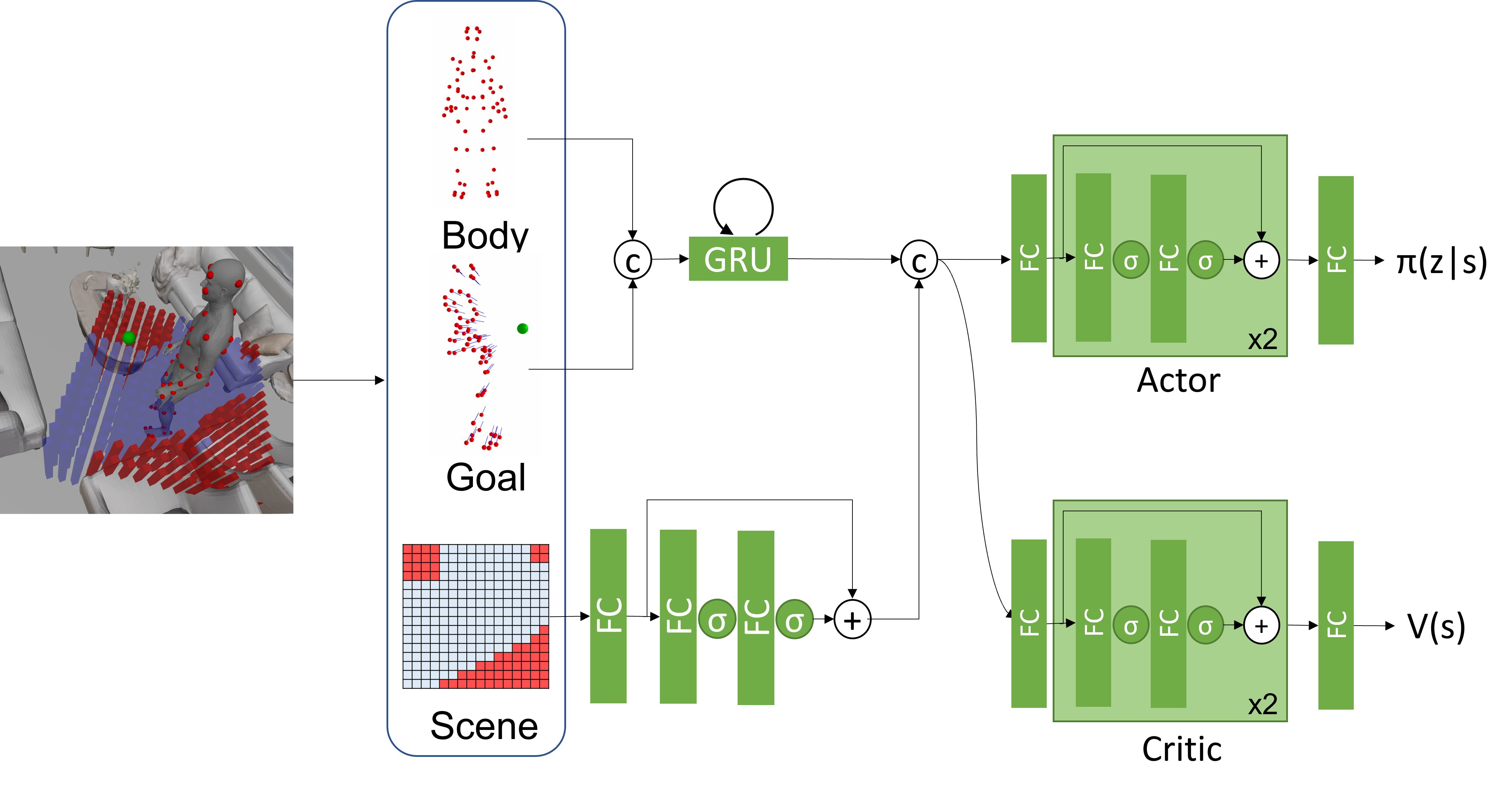}
\end{center}
\vspace{-0.5cm}
   \caption{Illustration of the scene-aware locomotion policy network. The locomotion policy state consists of the body markers, the goal-reaching feature of normalized direction vectors from markers to the goal pelvis, and the interaction feature of a 2D binary map indicating the walkability ( \textcolor{red}{red}: non-walkable area, \textcolor{blue}{blue}: walkable area) of the local $1.6m \times 1.6m$ square area. The locomotion policy network employs the actor-critic architecture and shares the state encoder.}

\label{fig:loco_policy}
\end{figure}

Navigating in cluttered scenes here means the human body moving to a target while avoiding collisions with scene objects.
Our key idea is to incorporate the walkability information of the surrounding environment into the states and use collision rewards to train the locomotion policy in order to avoid scene collisions.
Specifically, we represent the walkability of the environment surrounding the human agent using a 2D binary map $\mathcal{M} \in \{0, 1\}^{16 \times 16}$ as illustrated in Fig.~\ref{fig:loco_policy}. 

The walkability map is defined in the human's local coordinates and covers a $1.6m \times 1.6m$ area centered at the body pelvis and aligned with the body facing orientation. 
It consists of a $16 \times 16$ cell grid where each cell stores a binary value indicating whether this cell is walkable or not.
This local walkability map enables the policy to sense surrounding obstacles.

Referring to Eq.~\ref{equ:abstrac_state}, the person-scene interaction feature is specified by
\begin{equation}
    I = \text{vec}(\mathcal{M}),
\end{equation}
in which $\text{vec}(\cdot)$ denotes vectorization.
The goal-reaching feature is specified by
\begin{equation}
    G = (\tilde{\vec{g}}_p-\vec{X}_s)_n,
\end{equation}
where $\vec{X}_s$ and $ \tilde{\vec{g}}_p \in \mathbb{R}^{M \times 67 \times 3}$ are the body marker seed representing $M$ frames history of motion, and the broadcasted target pelvis location relative to the body-centered canonical coordinate, respectively. $(\tilde{\vec{g}}_p-\vec{X}_s)_n$ denotes the normalized vectors pointing from each marker to the goal pelvis.


The rewards contributing to Eq.~\ref{equ:abstract_reward} are defined as
\begin{equation}
r_{pene} = e^{-|\mathcal{M}_0 \cap \mathcal{B}_{xy}(X)|},
\label{equ:loco_pene_reward}
\end{equation}
where $\mathcal{M}_0$ denotes the non-walkable cells in the walkability map, $\mathcal{B}_{xy}(\cdot)$ denotes the 2D bounding box of the body markers $\vec{X}$, $\cap$ denotes their intersection, and $|\cdot|$ denotes the number of non-walkable cells overlapping with the human bounding box.

\begin{equation}
r_{goal} = r_{dist} + r_{ori},
\label{equ:loco_goal_reward}
\end{equation}
\begin{equation}
r_{dist} = 1-(\|\vec{p} -\vec{g}_p\|_2 - 0.05)_+,
\label{equ:loco_dist_reward}
\end{equation}
 
 \begin{equation}
r_{ori}=\frac{\langle \vec{o}, \vec{g}_p - \vec{p} \rangle}{2},
\label{equ:loco_ori_reward}
\end{equation}
where $r_{dist}$ encourages the body pelvis $\vec{p}$ to be close to the pelvis goal $\vec{g_p}$ and $r_{ori}$ encourages the body facing direction $\vec{o}$ to be aligned with the direction from  the current body pelvis $\vec{p}$ to the pelvis goal $\vec{g_p}$.




%% file: sections/method/interaction.tex
\subsection{Fine-grained Object Interaction Synthesis}
\label{sec:interaction}
To synthesize fine-grained human-object interactions, \eg sitting on a chair or lying on a sofa, we use body marker goals as guidance, and model the proximity between the body surface and the scene object in a compact way.
The goal marker sets can be generated by static person-scene interaction methods such as \cite{zhang_place_2020, hassan_populating_2021, zhao_compositional_2022}. 
We use COINS~\cite{zhao_compositional_2022} to generate the static goal interaction body for its performance and controllability of interaction semantics.

\begin{figure}[t]
\begin{center}
   \includegraphics[width=\linewidth]{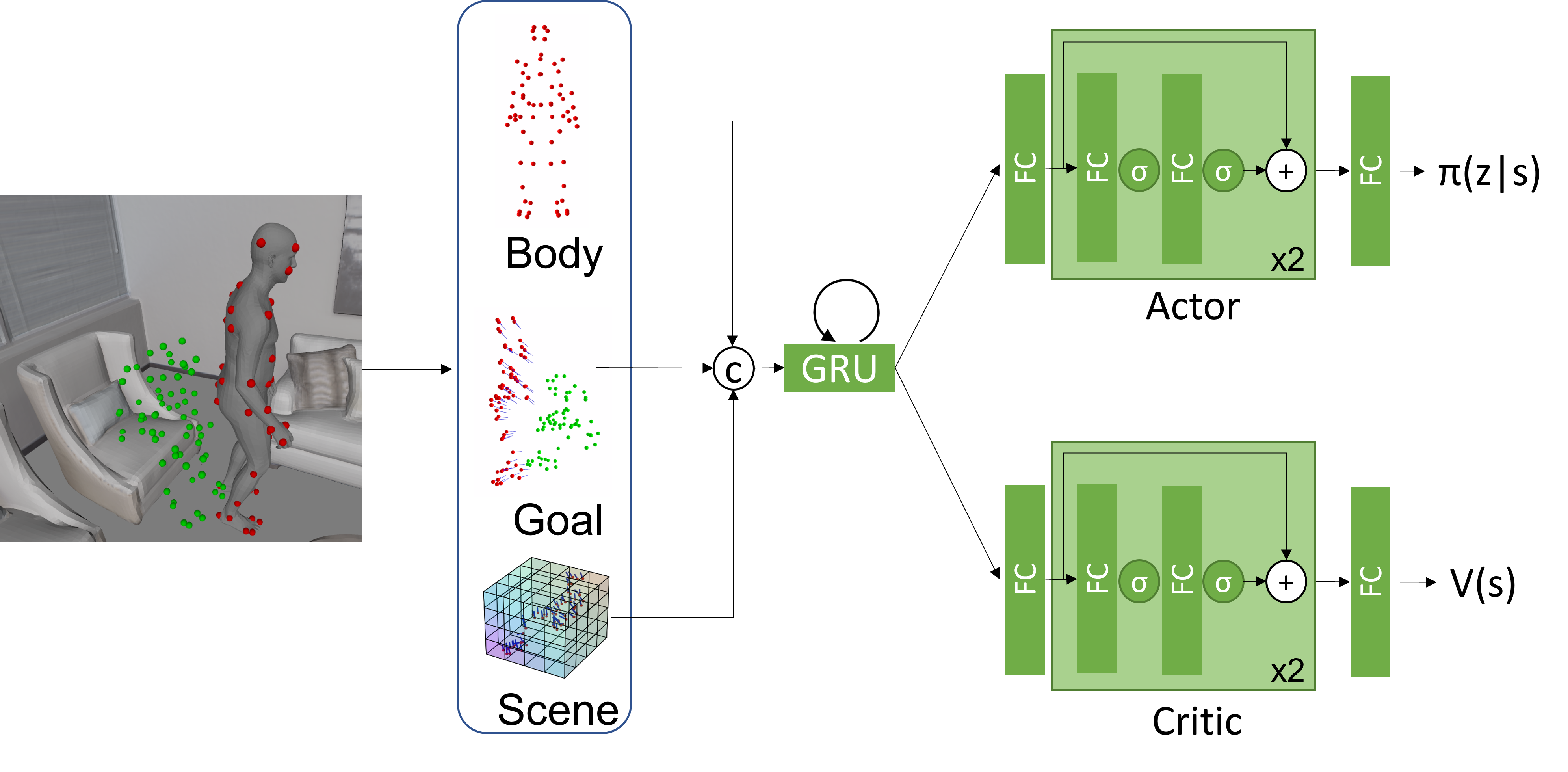}
\end{center}
\vspace{-0.5cm}
   \caption{Illustration of the object interaction policy network. The interaction policy state consists of the body markers, the goal-reaching features of both distance and direction from current markers to the goal markers, and the interaction features of the signed distances from each marker to the object surfaces and the signed distance gradient at each marker location.  Such interaction features encode the human-object proximity relationship.}
\label{fig:inter_policy}
\end{figure}

In addition to the marker-based goal guidance, we incorporate the proximity relations between humans and objects into the states.
Specifically, we use the signed distance from each marker to the surface of the scene object, as well as the gradient direction of the signed distance to represent the proximity relationship, as illustrated in Fig.~\ref{fig:inter_policy}. Both the signed distance and its gradient direction are calculated using the object's signed distance field (SDF) $\Psi_O$.

Referring to Eq.~\ref{equ:abstrac_state}, the person-scene interaction feature is formulated as
\begin{equation}
    I = [\Psi_O(\vec{X}_s), \nabla \Psi_O(\vec{X}_s)],
\end{equation}
in which $\Psi_O \in \mathbb{R}^{M \times 67}$ and $\nabla\Psi_O \in \mathbb{R}^{M \times 201} $ denote the SDF values and the gradient at each marker location in the $M$ frames, respectively, and $[\cdot,\cdot]$ denotes feature concatenation.
The goal-reaching feature is formulated as
\begin{equation}
    G = [(\tilde{\vec{g}}_m-\vec{X}_s)_n, \|\tilde{\vec{g}}_m - \vec{X}_s\|_2],
\end{equation}
in which $\vec{X}_s \in \mathbb{R}^{M \times 67 \times 3}$ denotes the body markers seed, $\tilde{\vec{g}}_m \in \mathbb{R}^{M \times 67 \times 3}$ denotes the broadcasted goal body markers, $(\tilde{\vec{g}}_m-\vec{X}_s)_n$ denotes the normalized vector representing the direction from each marker to the corresponding target marker, $\|\tilde{\vec{g}}_m - X_s\|_2$ denotes the distance from each marker to target marker.

%
 The interaction policy is trained using the reward defined in Eq.~\ref{equ:abstract_reward} with the following interaction-specific goal reward and penetration reward:
 
\begin{equation}
r_{goal} = 1-(\|\vec{x} -\vec{g_m}\|_2 - 0.05)_+,
\label{equ:inter_goal_reward}
\end{equation}

\begin{equation}
r_{pene} = e^{- \frac{1}{|V|}\frac{1}{T}\sum_{t=1}^{T} \sum_{i=1}^{|V|} |(\Psi_O(\vec{v_{ti}}))_-|},
\label{equ:inter_pene_reward}
\end{equation}
with $|V|$ being the SMPL-X mesh vertices, $T$ denotes the number of frames in each motion primitive (equals to 10 in our study).
The distance reward  encourages the final frame body markers $\vec{x}$ to be close to the goal body markers $\vec{g_m}$. 
The penetration reward penalizes all the body vertices within a motion primitive that have negative SDF values. We use body vertices instead of joints because human-object contact happens on the body surface and can be better detected using vertices.

Moreover, we train the interaction policy with a mixture of `sit/lie down' and `stand up' tasks. 
This training scheme enables the human agent to also learn how to stand up and transit from object interaction back to locomotion, which enables the synthesis of a sequence of interaction activities as in Fig.~\ref{fig:teaser}.

%% file: sections/experiment/experiment.tex
\section{Experiment}
\label{sec:experiment}

\paragraph{Motion Datasets.}
We combine the large-scale motion capture dataset AMASS \cite{mahmood_amass_2019} with SAMP \cite{hassan_stochastic_2021} motion data to train the motion primitive models.
%
Each sequence is first subsampled to 40 FPS and then split into 10 frames and 100 frames clips. Each motion clip is canonicalized using the first frame body.
Specifically, we select AMASS sequences annotated with `sit' or `lie' in BABEL \cite{punnakkal_babel_2021} and all motion data of SAMP to train the motion primitive model.
%
We train separate motion primitive models for locomotion and interactions using task-related data.
We observe that extending SAMP motion data with AMASS dataset is the key to making interaction policies work.

\paragraph{Policy Training Environments.}
To train the scene-aware locomotion policy, we randomly generate synthetic cluttered scenes consisting of random objects from ShapeNet \cite{chang_shapenet_2015}. 
%
Random initial and target location pairs are sampled in the walkable areas using navigation meshes to train the locomotion policy.
To train the interaction policy, we use the static person-object interaction data from PROX \cite{hassan_resolving_2019} and retargeted to ShapeNet objects.
%
We first randomly sample a furniture and interaction body goal from the retargeted PROX data. Then we sample the initial body with random poses and locations in front of the object to train the interaction policy. We also randomly swap the initial and goal body to learn both `sit/lie down' and `stand up' motions.
Please refer to Supp.~Mat. for more details.

\input{sections/experiment/locomotion}

\input{sections/experiment/interaction}

\input{sections/experiment/combination}

\subsection{Ablation Studies}
We perform ablation studies on the foot-ground contact reward and penetration avoidance reward of the interaction policies to substantiate the significance of these rewards.
We train two ablation interaction policies using identical networks and environments, differing only in the exclusion of one of these rewards in each case.
The quantitative metrics are reported in Tab.~\ref{tab:ablation}. 
The removal of the penetration avoidance reward yields a marked escalation in detected human-scene penetration. The removal of the foot-floor contact reward yields notably both inferior contact and penetration scores, accompanied by observed erratic synthesized motions that either remain suspended in the air or penetrate the floor. These empirical findings underscore the crucial role played by the foot-floor contact and penetration avoidance rewards in the learning of interaction policies.
\begin{table}[t]
\footnotesize
\centering
\caption{Rewards ablation studies results, where `-Contact' and `-Penetration' denote policies trained without the floor contact and penetration avoidance reward, respectively.}
    \begin{tabular}{lcccc}
    \hline
   & time $\downarrow$ & contact $\uparrow$ & pene. mean $\downarrow$ & pene. max $\downarrow$ \\
    \hline
    Ours & 4.09 & \textbf{0.97} & \textbf{1.91} & \textbf{10.61} \\
    - Contact & \textbf{3.75} & 0.88 & 19.49 & 60.10 \\
    - Penetration & 4.03 & 0.95 & 16.80 & 45.50 \\
    \hline
    \end{tabular}
    \label{tab:ablation}
\end{table}

%% file: sections/experiment/locomotion.tex
\subsection{Locomotion in 3D Scenes}
\label{sec:eval_loco}
We randomly generate test scenes for locomotion in the same way as the training scenes. 
The virtual human is instructed to move from the random start point to the random target point while avoiding penetration with scene objects.

\paragraph{Baselines and metrics.}
We compare our method with SAMP \cite{hassan_stochastic_2021} and GAMMA \cite{zhang_wanderings_2022} for locomotion. 
The SAMP results are recorded by running the released Unity demo. The start and termination are manually determined so the reported completion time may be slightly higher than the actual time due to human response time.
The evaluation metrics for locomotion include: 
1) time from start point to target point or reaching the time limit, measured in seconds. 
2) the average distance from the final body to the targets, measured in meters.
3) foot contact score encouraging the lowest feet joints on the floor and discouraging foot skating as defined in Eq.~\ref{equ:contact_reward}. We use body joints instead of markers to calculate the contact score because the marker set annotation for the SAMP body is missing.
%
4) locomotion penetration score indicating the percentage of body vertices that are inside the walkable areas according to the navigation mesh.



\begin{table}[t]
    \caption{Evaluation of the locomotion task. The
up/down arrows denote the score is the higher/lower the better and the best results are in boldface.
}
    \footnotesize
    \centering
    {
        \begin{tabular}{lcccccc}
        \hline
       & time $\downarrow$ & avg. dist $\downarrow$ & contact $\uparrow$ & loco pene $\uparrow$\\
        \hline
        SAMP \cite{hassan_stochastic_2021}  & 5.97 & 0.14 & 0.84 & 0.94\\
        GAMMA \cite{zhang_wanderings_2022} & \textbf{3.87} & \textbf{0.03} & 0.94 & 0.94\\
        Ours & 6.43 & 0.04 & \textbf{0.99} & \textbf{0.95}\\
        \hline
        \end{tabular}
    }
    \label{tab:metrics_loco}
\end{table}

\begin{figure}[t]
\begin{center}
   \includegraphics[width=0.45\linewidth]{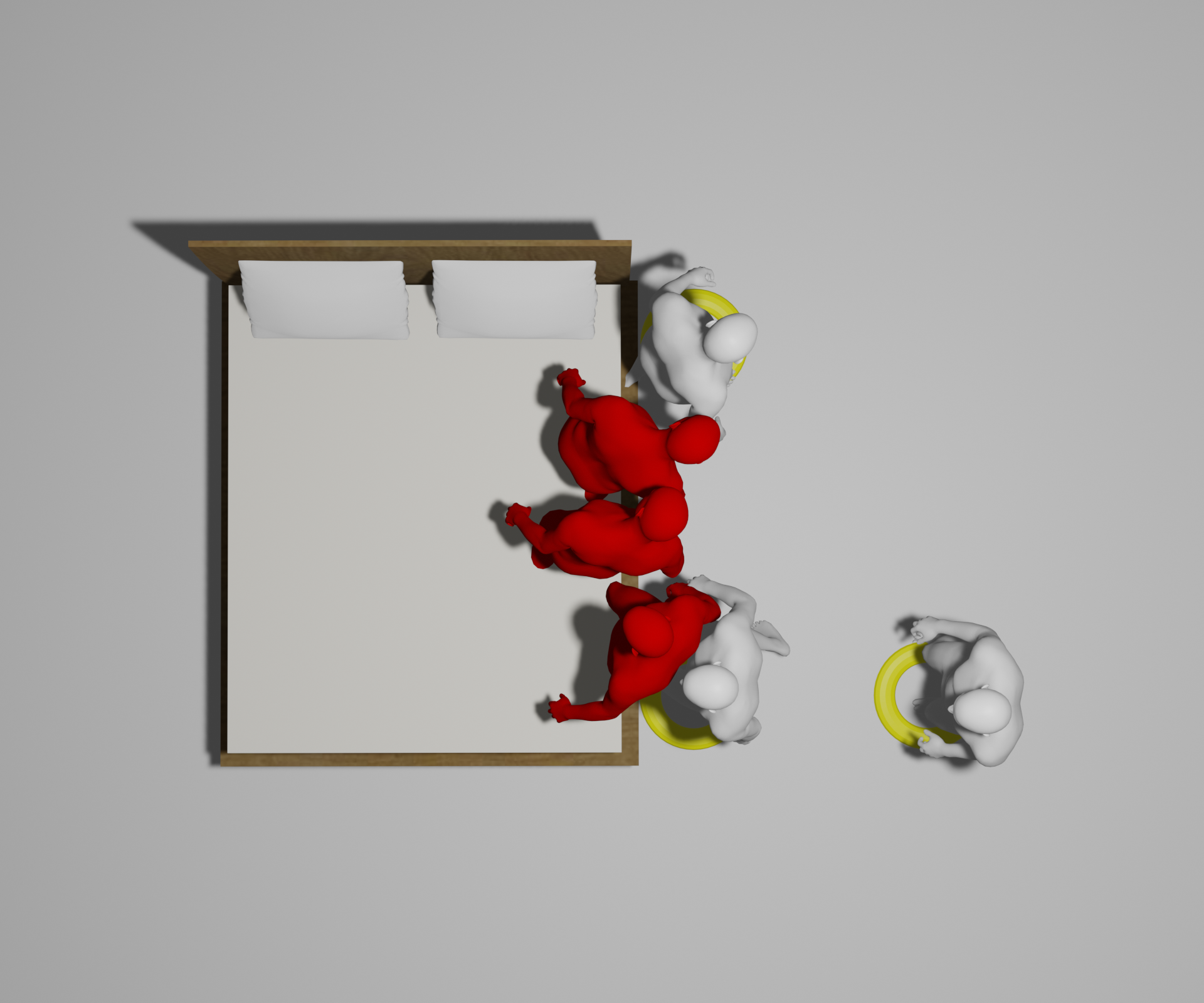}
   \includegraphics[width=0.45\linewidth]{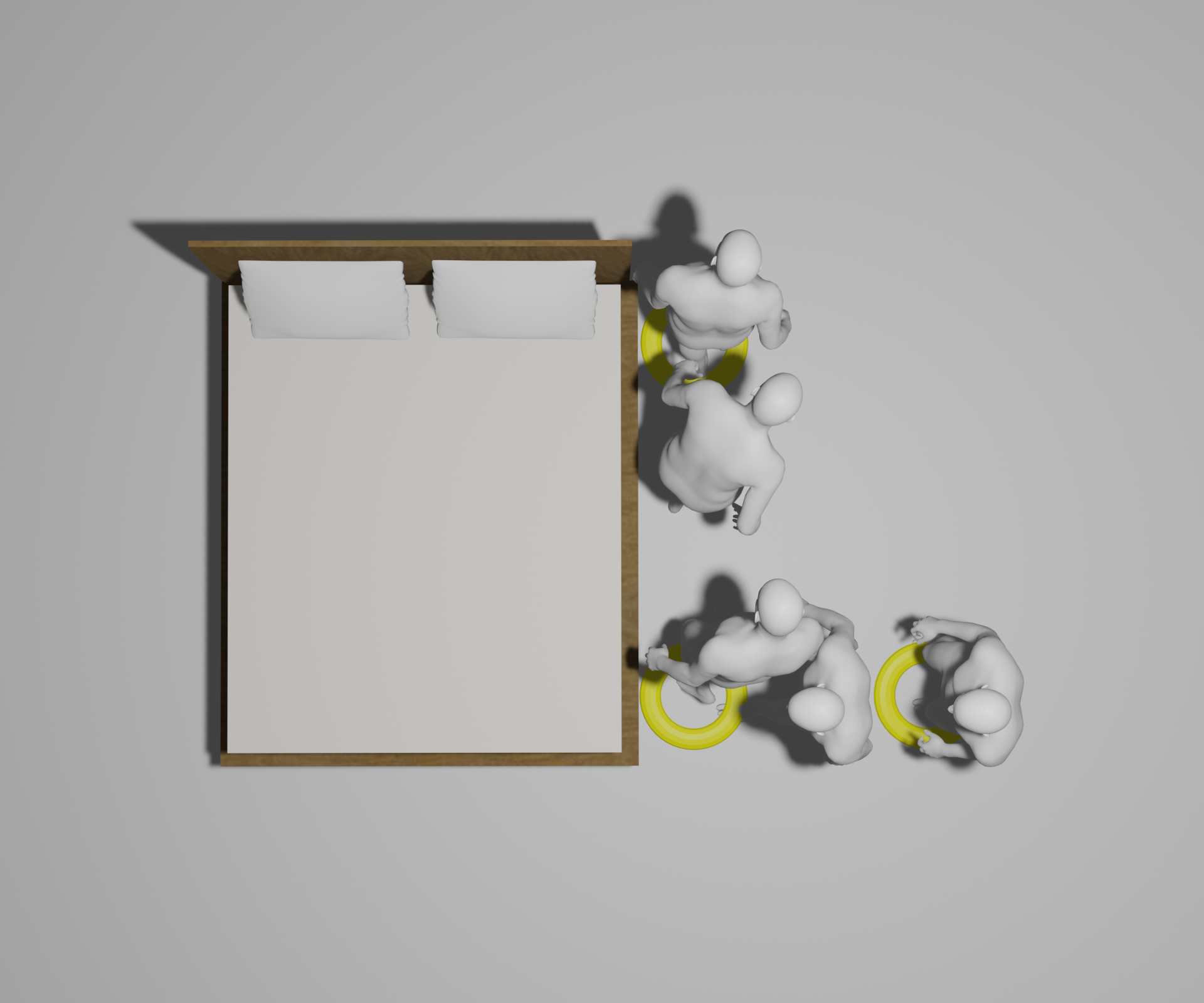}
   \includegraphics[width=0.45\linewidth]{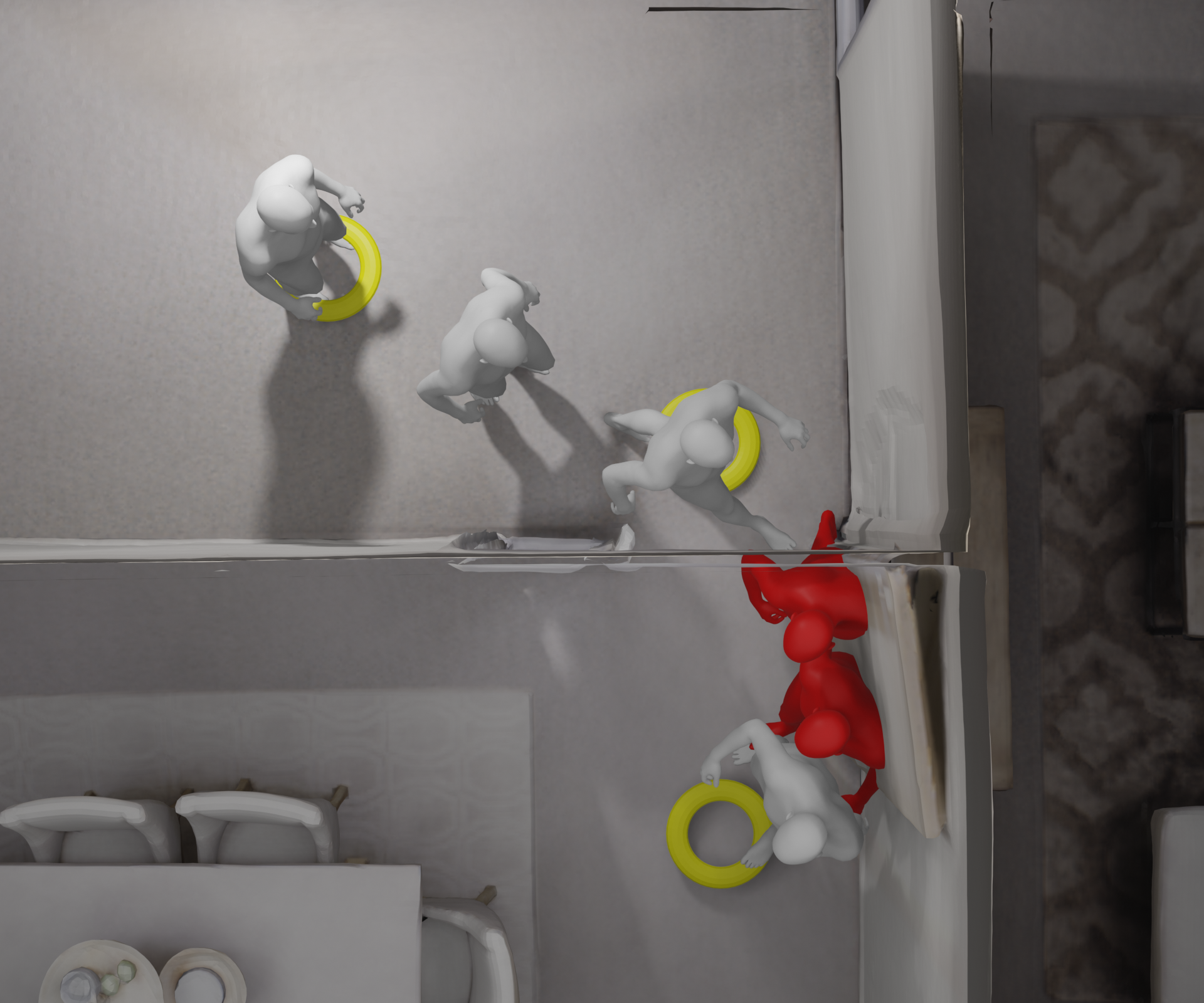}
   \includegraphics[width=0.45\linewidth]{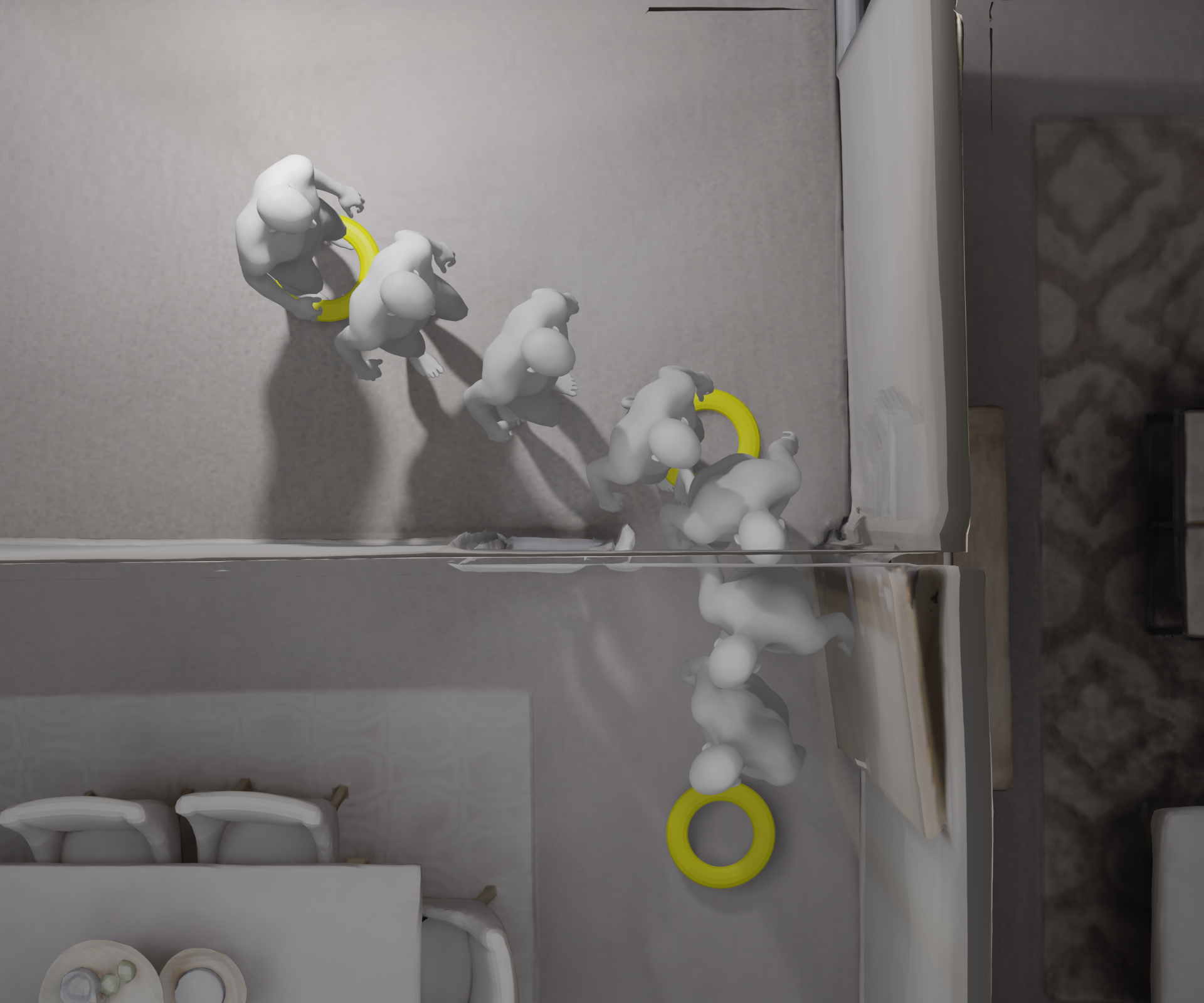}
\end{center}
\vspace{-5mm}
   \caption{Demonstration of locomotion tasks where GAMMA \cite{zhang_wanderings_2022} (left) collides with the obstacles (\textcolor{red}{red} bodies) while our scene-aware locomotion policy (right) avoids collision.  The yellow circles denote the specified waypoints.}
\label{fig:loco-demo}
\end{figure}

\paragraph{Results.}
Table \ref{tab:metrics_loco} shows the empirical evaluation results. 
Our method achieves a higher contact score (0.99) than both GAMMA (0.94) and SAMP (0.84) which indicates better foot-floor contact and less foot skating. 
Moreover, our method achieves the highest penetration score which indicates our scene-aware policy can better avoid scene collisions.
Fig.~\ref{fig:loco-demo} shows examples of locomotion tasks where GAMMA collides into the scenes while our scene-aware policy successfully avoids penetration.
Compared to GAMMA trained in the same environments, we observe our scene-aware policy learns more conservative behavior with lower moving speed and spent more time (6.43s) walking to the target locations, just like a human afraid of stepping on surrounding traps.
All the methods can reach the target location within a reasonable distance.

%% file: sections/experiment/interaction.tex
\subsection{Fine-Grained Human-Object Interaction}
\label{sec:eval_inter}
We evaluate the object interaction task on 10 unseen objects (3 armchairs, 3 straight chairs, 3 sofas, 1 L-sofa) from ShapeNet \cite{chang_shapenet_2015}.  
We use the object size annotation of ShapeNet and manually filter unrealistic-sized objects.
The virtual human is randomly placed in front of the target object and then instructed to perform the interaction, stay for around 2 seconds, and then stand up.
We evaluate two interactions of sitting and lying separately.

\paragraph{Baselines and metrics.}
We compare our method with SAMP \cite{hassan_stochastic_2021} for the object interaction task.
The evaluation metrics for the interaction tasks are:
1) time of completing the object interaction task.
2) foot contact score as defined in Eq.~\ref{equ:contact_reward}. Note that the foot contact score does not always reflect the motion quality for lying tasks because the foot can often be off the floor during lying.
3) interaction penetration score for each frame is defined as:

\begin{equation}
s_{inter\_pene} = \sum_{v_i \in V} |(\Psi_O(v_i))_-|,
\label{equ:inter_pene_score}
\end{equation}

where $\Psi_O$ is the object signed distance field, $(\cdot)_-$ clips all positve distance values to zero, and $V$ is the body vertivces. 
We show both the average penetration over time and the maximum penetration in one sequence.
%
%

\begin{table}[t]
    \caption{Evaluation of the interaction tasks. The
up/down arrows denote the score is the higher/lower the better and the best results are in boldface.}
    \footnotesize
    \centering
    \resizebox{\columnwidth}{!}
    {
        \begin{tabular}{lccccc}
        \hline
       & time $\downarrow$ & contact $\uparrow$ & pene. mean $\downarrow$ & pene. max $\downarrow$ \\
        \hline
        SAMP \cite{hassan_stochastic_2021} sit & 8.63 & 0.89 & 11.91  & 45.22\\
        Ours sit & \textbf{4.09} & \textbf{0.97}  & \textbf{1.91} & \textbf{10.61}\\
        \hline
        SAMP \cite{hassan_stochastic_2021} lie & 12.55 & 0.73 & 44.77 & 238.81\\
        Ours lie & \textbf{4.20} & \textbf{0.78} & \textbf{9.90}  & \textbf{44.61}\\
        \hline
        \end{tabular}
    }
    \label{tab:metrics_inter}
\end{table}
        
\begin{figure}[t]
\begin{center}
\includegraphics[width=\linewidth]{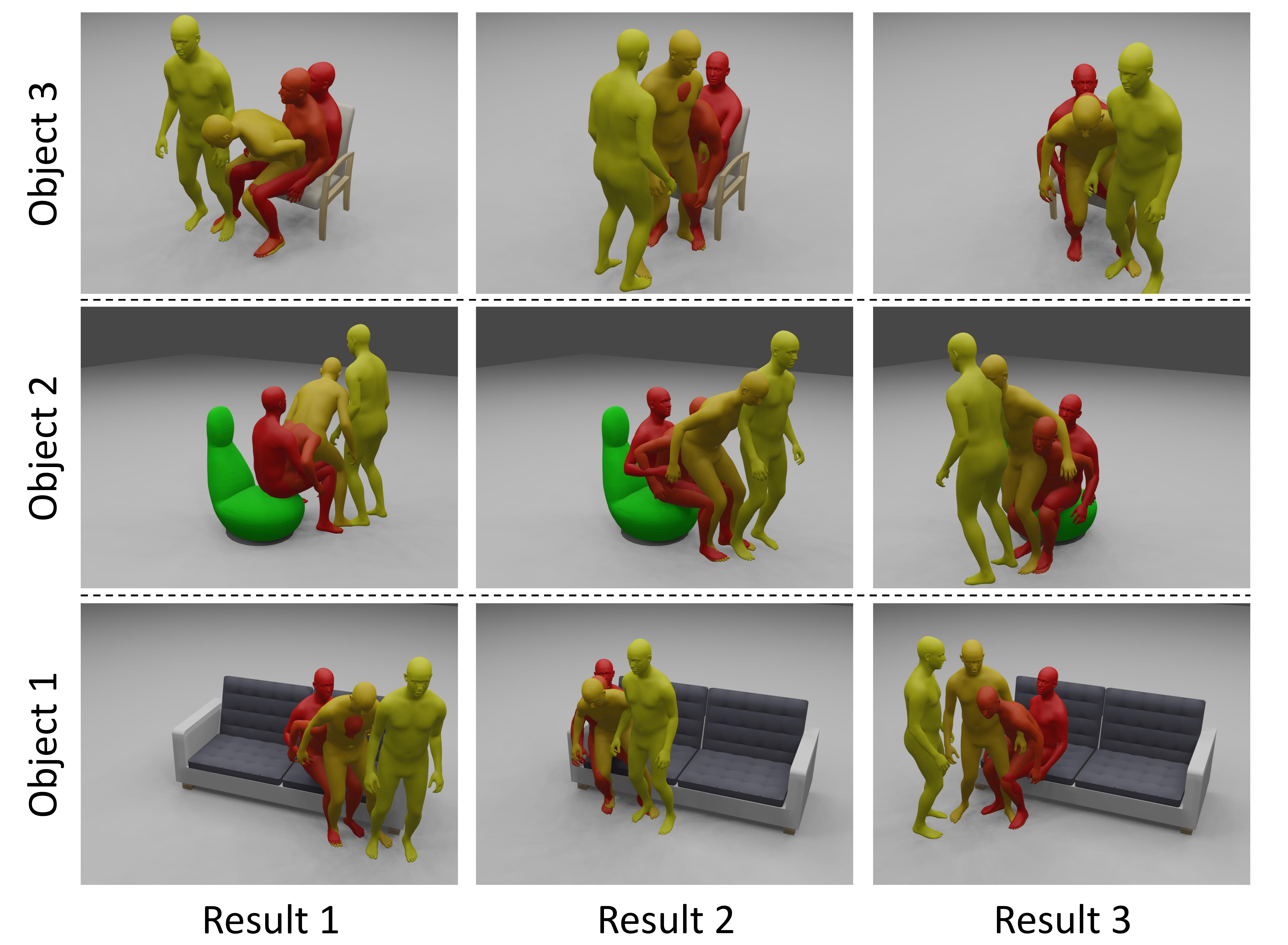}
\vspace{-8mm}
\end{center}

   \caption{Demonstration of various generated object interactions. Each row shows generated interactions with the same object starting from random initial body locations and orientations. Colors from \textcolor{darkyellow}{yellow} to \textcolor{red}{red} denote time.}
\label{fig:inter-demo}
\end{figure}

\paragraph{Results.}
Tab.~\ref{tab:metrics_inter} shows the evaluation results. 
Our method achieves a significantly higher foot contact score, indicating more natural motion. 
Our method also achieves lower mean and maximum penetration, which means our method generates more physically plausible results. 
Moreover, our method can complete the interaction tasks much faster than SAMP. This is because SAMP does not generalize well to unseen random body initialization and needs a longer time to start performing interactions. 
%
Qualitative results demonstrating the various object interactions generated by our method are shown in Fig.~\ref{fig:inter-demo}. Our object interaction policy generalizes to random initial body locations and orientations, as well as novel objects of various shapes.

%% file: sections/experiment/combination.tex
\subsection{Interaction Sequences in 3D Scenes}
\begin{figure}[t]
    \includegraphics[width=\linewidth]{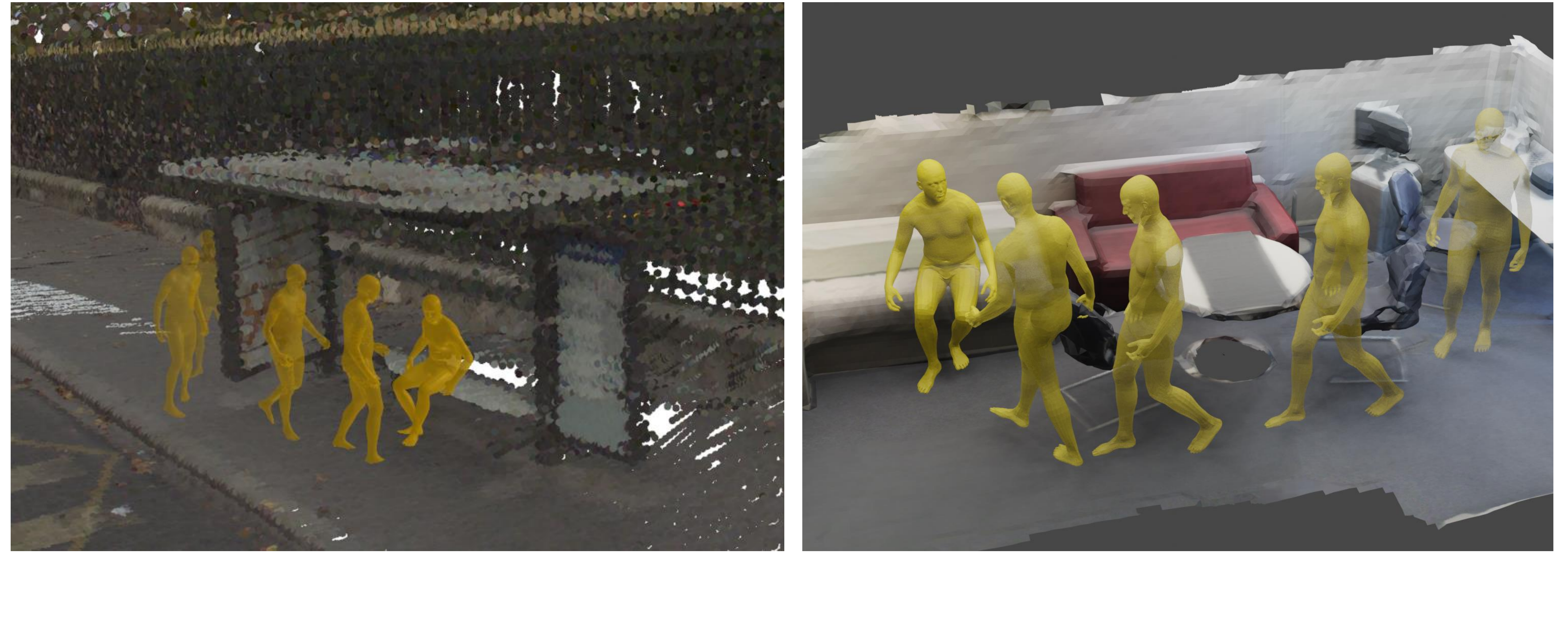}
    \vspace{-10mm}
    \caption*{\parbox{\linewidth}{\small
    \hspace{3mm} Walk to sit on the bus bench  \hspace{8mm} Walk to sit on the bed \hspace{5mm}}
    }
    \caption{Our method can synthesize natural human movements in scene reconstructions that are significantly different from the training scenes. We show the synthesis results in one Paris street \cite{deschaud2021pariscarla3d} (left figure) and one indoor room \cite{hassan_resolving_2019} (right figure). Transparent to solid colors denote time.}
    \label{fig:generalize}
\end{figure}

\label{sec:inter_seq}
Humans often continuously perform sequences of interactions with various objects in complex real-world scenes  as shown in Fig.~\ref{fig:teaser}. Such interaction sequences are combinations of alternating locomotions and fine-grained interactions which we evaluate in Sec.~\ref{sec:eval_loco} and Sec.~\ref{sec:eval_inter} respectively. 
We conduct the empirical evaluation in real scene scans from Replica \cite{straub_replica_2019} and compare our method with SAMP.
We select a list of interactable objects in the scene and instruct the virtual human to walk to the objects to perform interactions one by one. 
The evaluation metrics include the foot contact score, locomotion penetration score, and the mean and maximum interaction penetration score. 
We also conducted a perceptual study where participants are shown a side-by-side comparison of results from two methods and asked to choose the one perceptually more natural. We report the rate of being chosen as the better.


\begin{table}[t]
    \caption{Evaluation of interaction sequences synthesis.  Here `contact' denotes the foot contact score, `loco. pene.' is the percentage of body vertices inside the walkable areas, `inter. pene. mean/max' denotes the mean and maximum penetration with interaction objects, and `perceptual' denotes the ratio of being chosen as perceptually more natural. The best results are in boldface.}
    \centering
    \small
    \begin{tabular}{lcc}
        \toprule
       & SAMP~\cite{hassan_stochastic_2021} & Ours \\
       \midrule
       contact $\uparrow$ & 0.87 & \textbf{0.96}\\
       loco. pene. $\uparrow$ & 0.62 & \textbf{0.72} \\
       inter. pene. mean$\downarrow$ & 15.61 & \textbf{3.40} \\
       inter. pene. max $\downarrow$ & 101.25 & \textbf{39.68} \\
       perceptual. $\uparrow$ & 0.15 & \textbf{0.85} \\
       \bottomrule
    \end{tabular}
    \label{tab:metrics_combined}
\end{table}

The evaluation results are shown in Tab.~\ref{tab:metrics_combined}. 
Our method generates high-quality results of human-scene interactions in cluttered environments. 
Our results achieve higher contact scores and less penetration with the scenes compared to SAMP.
In addition, our method generates perceptually more natural results according to perceptual study.

%% file: sections/conclusion.tex
\section{Discussion and Conclusion}

\paragraph{Limitations and Future Work.}
Our current method has various limitations that could be improved in future works.
First, our method does not fully resolve penetrations with scene objects and floors since our method only uses rewards to encourage avoiding penetration with scenes, which does not impose hard constraints for penetrations. The combination of our method with physics simulation holds the potential to effectively address and resolve penetration issues.
Moreover, the lying motions generated by our method are not as natural as the sitting motions because the motion primitive model fails to learn a comprehensive action space for lying given limited training data. Specifically, the available AMASS motion capture data for lying (167 seconds) is significantly less than the data for sitting (5K seconds). To overcome this issue, we aim to explore more data-efficient learning methods and scalable methods to collect human-scene interaction data.
Furthermore, our locomotion policy is now limited to flat-floor scenes due to its reliance on the 2D occupancy-based walkability map. However, to broaden the applicability of our approach to more complex scenes such as uneven outdoor terrains and multi-floor buildings requiring walking upstairs, it will be necessary to replace the walkability map with a more suitable environment sensing mechanism.
In addition, our method is restricted to interactions with static scenes. However, in the real world, humans are exposed to dynamic interactions with movable objects and scenes involving autonomous agents including other humans, animals, and vehicles. Extension to dynamic interactions will enable tackling a broader range of intricate and dynamic applications that mirror the complexities inherent in the real world.


\begin{figure}[t]
    \vspace{-5mm}
    \includegraphics[width=\linewidth]{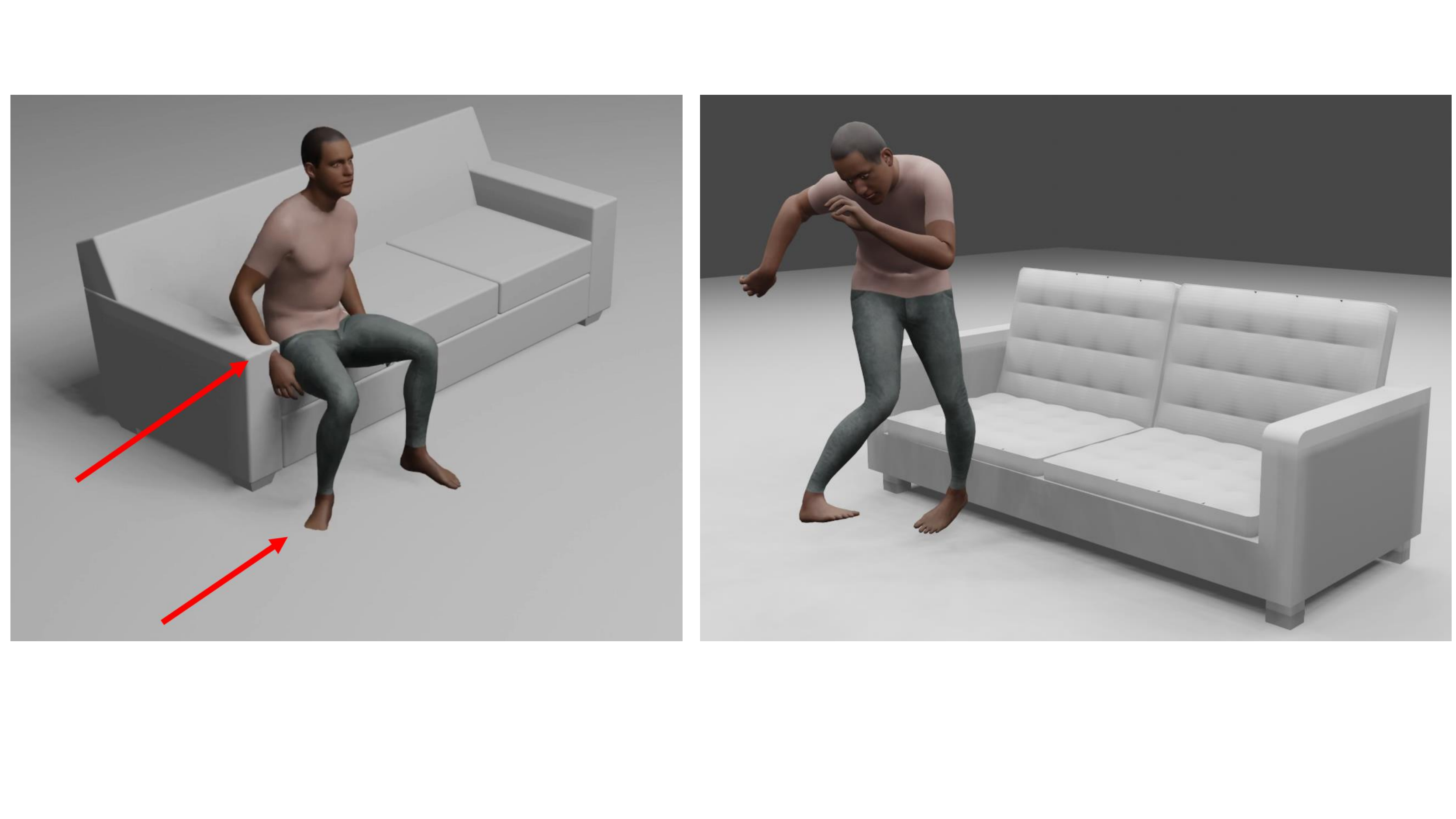}
    \vspace{-16mm}
    \caption*{\parbox{\linewidth}{\hspace{6mm} Inter-penetrations \hspace{18mm} Unnatural poses}}
   \caption{Limitations. Penetrations remain observable in our results (left) since our method only encourages avoiding collision with rewards. Furthermore, the absence of sufficient training data for lying motions impedes the motion primitive model's ability to acquire a comprehensive lying motion space, leading to degraded motion (right).}
\label{fig:limit}
\end{figure}

\paragraph{Conclusion.}
In this paper, we leverage reinforcement learning to establish a framework to synthesize diverse human motions in indoor scenes, which is stochastic, realistic, and perpetual. The proposed method has large potential to improve many applications such as daily-living activity simulation, synthetic data creation, architecture design, and so on.
Compared to existing methods, our method realizes fine-grained control by using body surface keypoints as targets, and achieves autonomous body-scene collision avoidance by incorporating scene information into the states and the rewards. 
Experiments show that our method effectively enables virtual humans to inhabit the virtual, and outperforms baselines consistently.
%

\paragraph{Acknowledgements.} 
We sincerely acknowledge the anonymous reviewers for their insightful suggestions. 
This work was supported by the SNSF project grant 200021 204840 and SDSC PhD fellowship.
Lingchen Yang provided valuable suggestions on visualization.

\clearpage

%% file: sections/supplementary.tex
\begingroup
\onecolumn 
\clearpage

\appendix

\begin{center}
\Large{\bf Synthesizing Diverse Human Motions in 3D Indoor Scenes \\
\vspace{0.3cm} 
**Appendix**\\
}
\vspace{0.2cm}
\vspace{1cm}
\end{center}

\setcounter{page}{1}
\setcounter{table}{0}
\setcounter{figure}{0}
\renewcommand{\thetable}{S\arabic{table}}
\renewcommand\thefigure{S\arabic{figure}}

\section{Policy Training Environment Details}
\subsection{Locomotion Environments}
We train the scene-aware locomotion policy using random synthetic scenes to learn generalizable locomotion skills of moving from the initial location to the goal location while avoiding collision with scenes. The initial and goal locations used for training are waypoints of collision-free paths sampled in the synthetic scenes.

Each synthetic scene has a random scene size, consists of random numbers and categories of objects sampled from ShapeNet \cite{chang_shapenet_2015}, and has a random scene layout.
The synthetic scenes are generated using the following steps: 
\begin{itemize}
\item  Sample the initial scene shape as a rectangle with edges ranging from 2 meters to 7 meters.
\item Randomly sample furniture objects constituting the scene from ShapeNet. Specifically, we sample objects from chairs, beds, sofas, desks, and tables. We limit the number of objects belonging to categories that normally have large sizes (e.g.~beds) to avoid the scenes being fully occupied, leaving no space for human movements. We use the real object size annotation of ShapNet and transform the object model to make the z-axis point up.
\item Randomly rotate and translate the objects in the scene to obtain random scene layouts.
\item Expand the scene boundary so that every object keeps a reasonable distance from the boundary and humans can potentially walk by.
\end{itemize}

After synthetic scene generation, we calculate the corresponding navigation mesh as described in Sec.~\ref{sec:navmesh}, and randomly sample pairs of collision-free initial-goal locations in the walkable areas. We first randomly sample two initial and goal locations on the navigation mesh. Then we use navigation mesh-based pathfinding to generate a sequence of waypoints that constitute a collision-free path. Each pair of consecutive waypoints are used as one initial-goal location pairs to train the locomotion policy.
One sample synthetic scene and waypoints for training the locomotion policy are shown in Fig.~\ref{fig:env_loco}.

We train the locomotion policy using the synthetic scenes and corresponding initial-goal location pairs. The locomotion policy is trained to move from the initial location to the goal location while avoiding penetration with scene objects. We further randomize the initial body pose and orientation to make the policy generalize to various initial body configurations.

\begin{figure}[ht]
\begin{center}
   \includegraphics[width=0.6\linewidth]{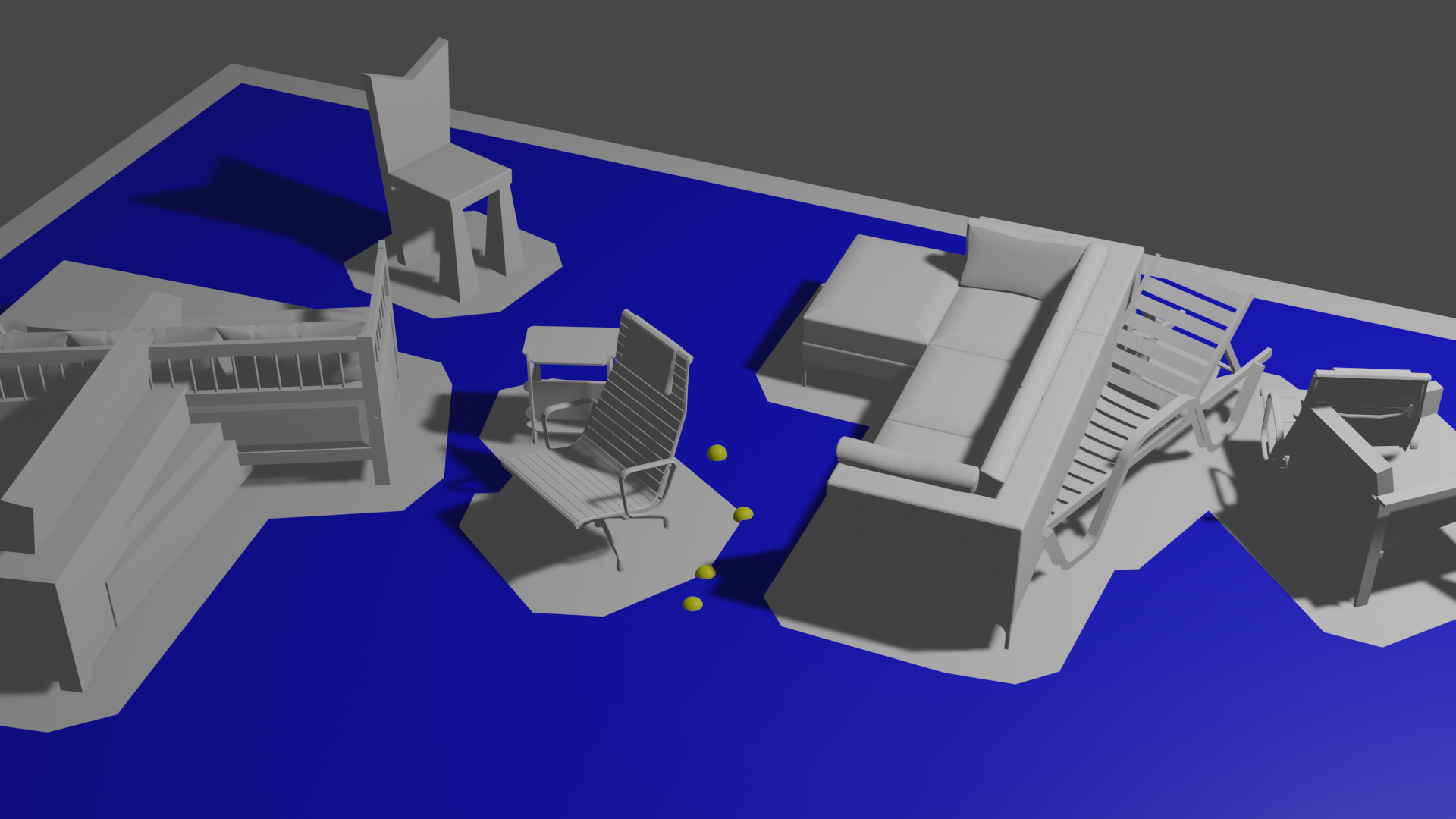}
\end{center}
   \caption{Illustration of a synthetic scene and sampled waypoints used to train the locomotion policy. The corresponding navigation mesh is visualized as blue, denoting the walkable areas for humans. The yellow spheres denote pair-wise collision-free waypoints found by navigation mesh-based path finding and are used to train the locomotion policy. }

\label{fig:env_loco}
\end{figure}

\subsection{Object Interaction Environments}
We train the human-object interaction policy to reach the fine-grained body marker goals that perform the specified interaction. 
The static goal human-scene interaction data is the prerequisite for training the interaction policy, which we obtain from the PROX \cite{hassan_resolving_2019} dataset using the following steps:

\begin{itemize}
\item We obtain the static human-object interaction estimation from PROX recordings, which consist of SMPL-X body estimation from LEMO \cite{Zhang:ICCV:2021}, and object mesh from the instance segmentation and annotation from COINS \cite{zhao_compositional_2022}. Specifically, we use the static human-scene interactions annotated as `sit on' and `lie on' according to COINS.

\item To improve object diversity and augment the interaction data from PROX, we retarget the static interaction data to random ShapeNet \cite{chang_shapenet_2015} objects similar to the data augmentation in \cite{starke_neural_2019}. For each static human-object interaction data from PROX,  we randomly sample an object from ShapeNet and fit it to the original PROX object by optimizing scale, rotation, and translation. 
After fitting, we replace the original PROX object with the fitted object. Then we augment the interaction data by applying slight scaling and rotation augmentation to the fitted object, and the corresponding human bodies are updated using contact points and relative vectors similar to \cite{starke_neural_2019}. 
\end{itemize}

With the object retargeting and augmentation, we obtain goal static human-object interaction data with increased diversity compared to the original PROX dataset.
When training the object interaction policy, we randomly sample one frame of static interaction to retrieve the interaction object mesh and fine-grained goal body markers for the training environment setup.  We randomly sample the initial body location and pose in front of the interaction object. Furthermore, we randomly swap the initial body and goal body with a probability of 0.25, in order to also learn `stand up' behaviors in addition to `sit/ lie down' behaviors.
Two example training scenes of sitting down and standing up are demonstrated in Fig.~\ref{fig:env_inter}.

\begin{figure}[t]
\begin{subfigure}[b]{0.48\linewidth}
    \includegraphics[width=\linewidth]{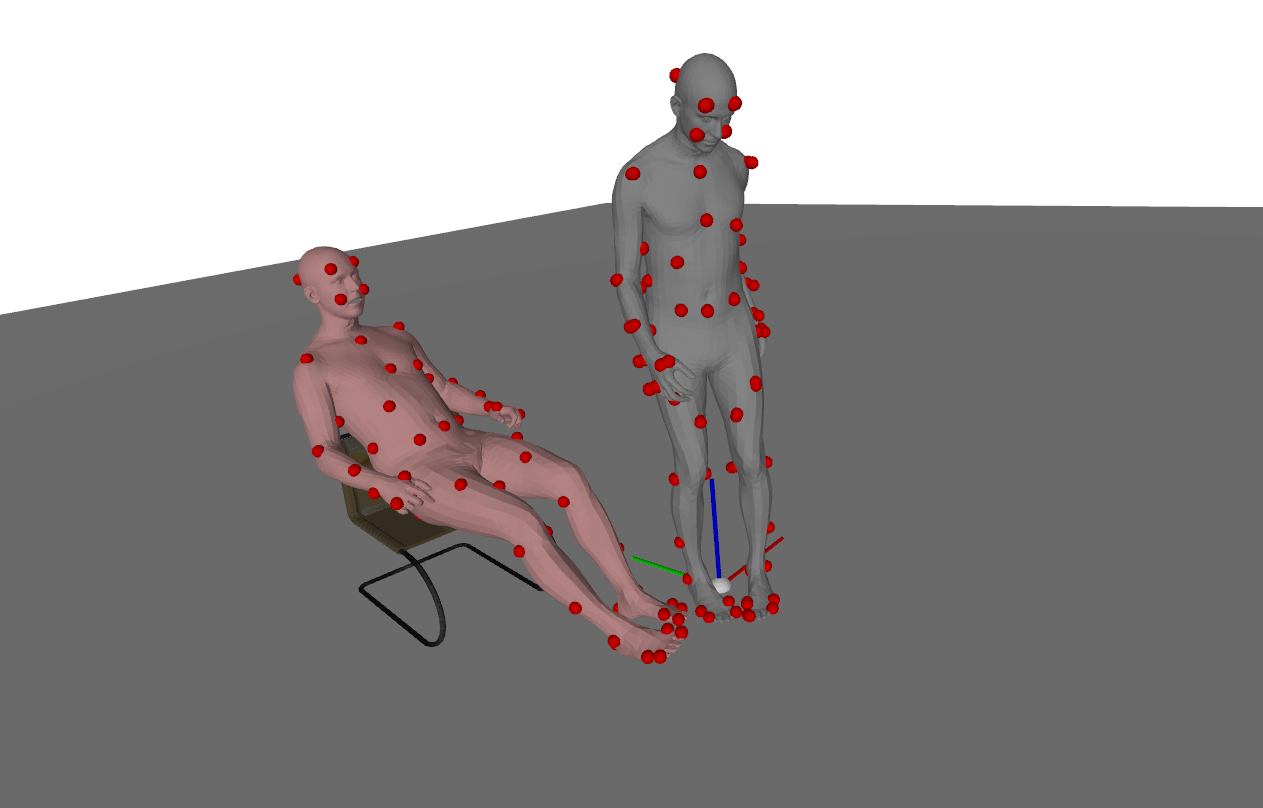}
    \caption{`Sit down' training environment}
    \label{fig:subfig1}
  \end{subfigure}
  \hfill
  \begin{subfigure}[b]{0.48\linewidth}
    \includegraphics[width=\linewidth]{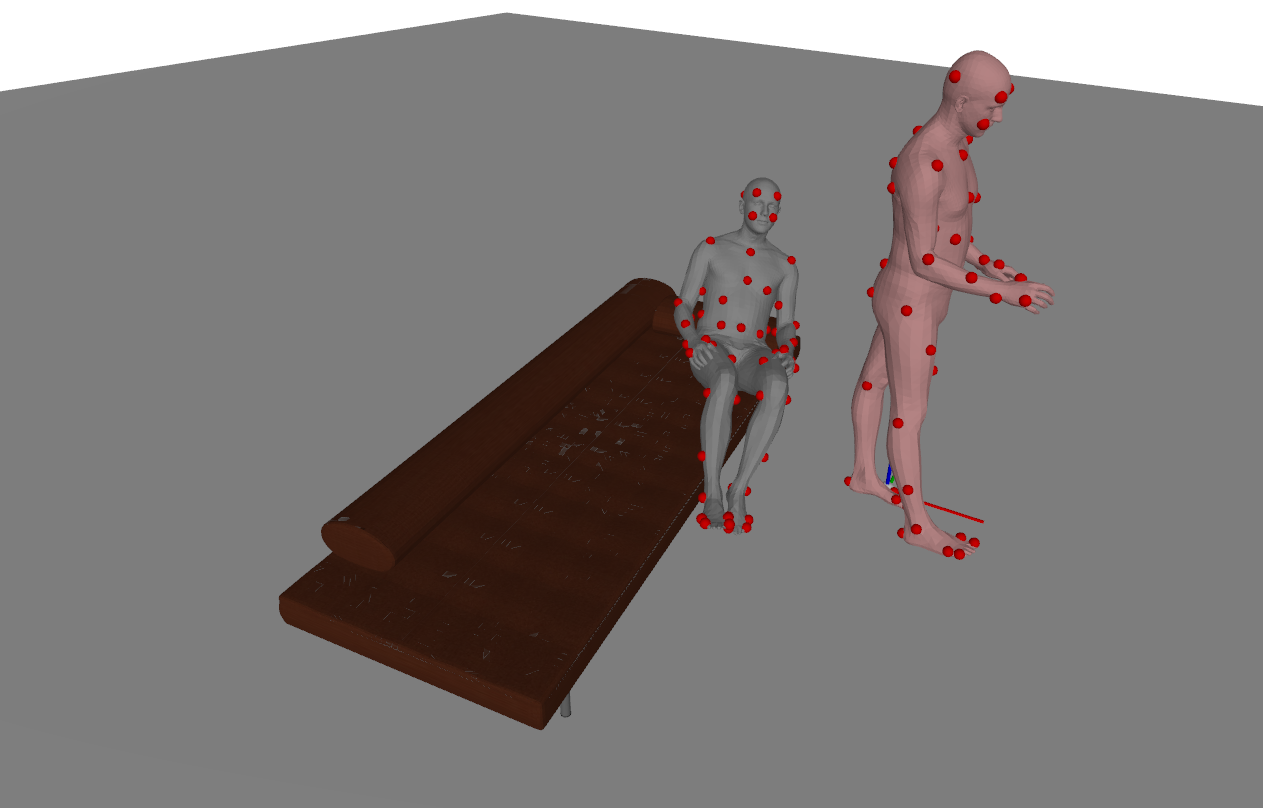}
    \caption{`Stand up' training environment}
    \label{fig:subfig2}
  \end{subfigure}
   \caption{Demonstration of two example training environments (left: sit down, right: stand up) for the fine-grained human-object interaction policy. Each training scene consists of an interaction object, an initial body (gray), and a goal body (pink). The object interaction policy is trained to reach the goal interaction body while avoiding collision with the interaction object and the floor. The red spheres denote the body markers.}
\label{fig:env_inter}
\end{figure}

\section{Implementation Details}
\paragraph{Goal static interaction synthesis using COINS.}
In this paper, we use a modified version of COINS\cite{zhao_compositional_2022}, incorporating slightly improved object generalization, for the synthesis of goal static human-scene interactions.
COINS synthesizes static human-scene interactions conditioned on interaction semantics and object geometries. However, the original COINS models are trained on the PROX \cite{hassan_resolving_2019} dataset which contains a very limited number of object models. 
This restricted training object diversity constrains the object generalization capabilities of COINS models. Empirical observations reveal that generation quality deteriorates when applied to objects with domain gaps, such as CAD models from ShapeNet \cite{chang_shapenet_2015} and in-the-wild scans from ScanNet \cite{dai2017scannet}. We observed that the object generalization failures are mainly due to the pelvis generation stage. Therefore, we annotated the pelvis frame data of sitting and lying interactions on a subset of ShapeNet objects that covers more diverse objects than contained in the PROX dataset. We retrain the PelvisNet of COINS with the annotated pelvis frame data and keep the BodyNet and other sampling algorithms untouched. 
The goal static interactions are generated and filtered in advance of the interaction motion synthesis.

\paragraph{Navigation mesh and path planning.}
\label{sec:navmesh}
We use the open-sourced pynavmesh \cite{itrch_pynavmesh} library for navigation mesh creation and collision-free pathfinding.
We implemented utility code to adapt the library to general scenes with the z-axis pointing up.
We enforce the generated navigation mesh only containing triangle faces. For each scene, we create two versions of navigation meshes with different agent radii. The navigation mesh created with a larger agent radius of 0.2 is used for collision-free pathfinding, and the other one created with a smaller radius of 0.02 gives a tight fitting of the unoccupied areas and is used for the local walkability map calculation, as described in the next paragraph.

\paragraph{Walkability map.}
The walkability map is implemented as a 2D occupancy map centered at the human pelvis and aligns with the body's forward orientation. 
We leverage a 16x16 walkability map covering 1.6 meters by 1.6 meters square area. Each cell of the map has a binary value indicating whether this cell is walkable or occupied by obstacles.
At each time step, we dynamically update this human-centric local walkability map. We first sample the 256 cells in the human-centric local coordinates frames, then transform the cell center to the scene coordinates, and evaluate each cell occupancy using the tightly fitted navigation mesh by querying whether the centroid is inside any triangle faces of the navigation mesh.
One example walkability map is shown in Fig.~\ref{fig:walk_map}.

\begin{figure}[t]
\begin{center}
   \includegraphics[width=0.3\linewidth]{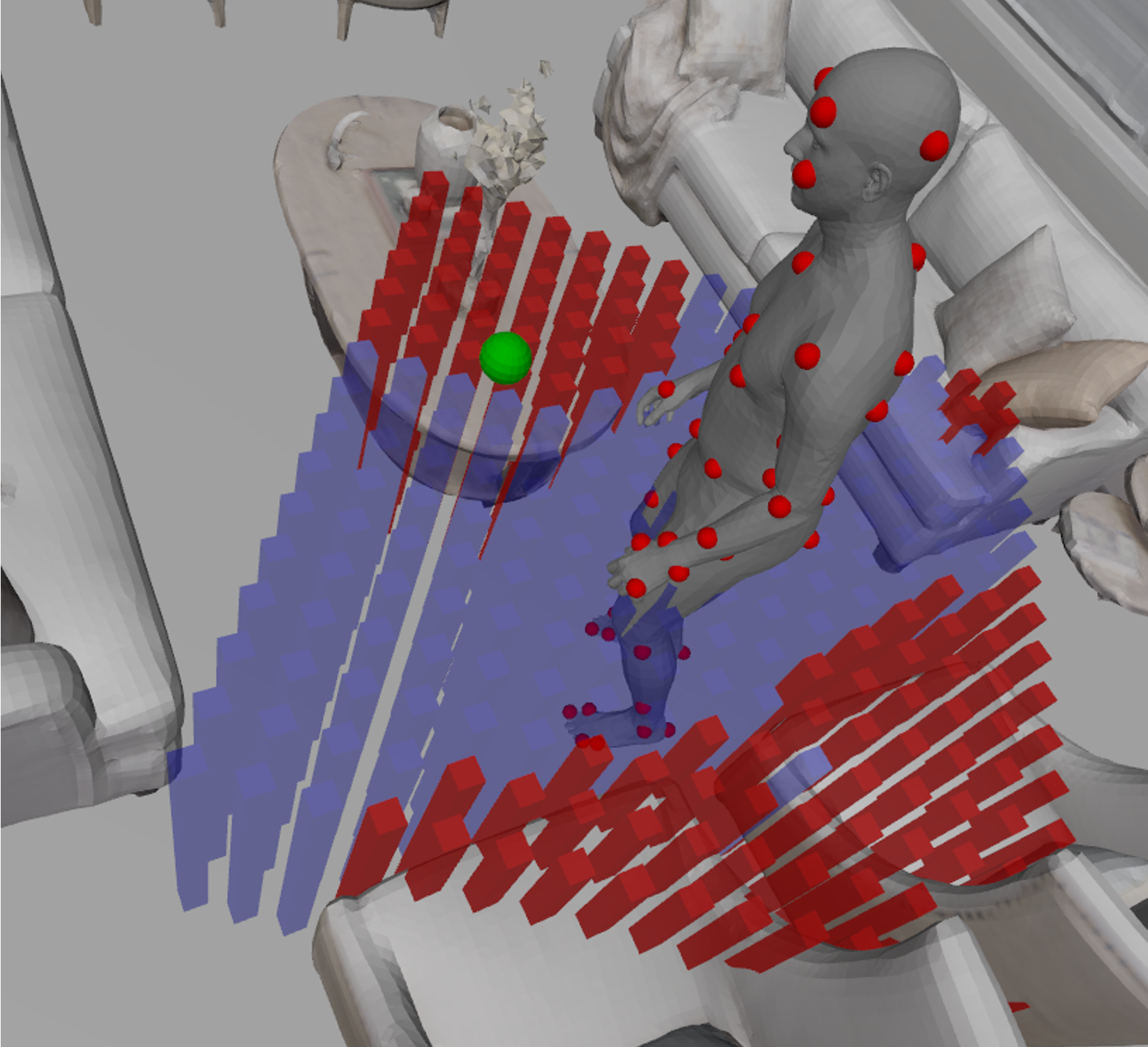}
\end{center}
   \caption{Illustration of the human-centric walkability map. The walkability map is a 2D occupancy map indicating which areas surrounding the human are walkable (blue cells) or occupied by obstacles (red cells). The walkability map is dynamically updated at each step according to the human body's location and orientation. }

\label{fig:walk_map}
\end{figure}

\paragraph{SDF-based features.}
We leverage the mesh-to-sdf \cite{kleineberg_mesh_to_sdf} library to calculate the marker-object signed distance and gradient features, which are used by the fine-grained human-object interaction policy. 
For each interaction object, we precompute a 128x128x128 SDF grid and a corresponding gradient grid. At each test step of interaction synthesis, we calculate the body marker-object distance and gradient features by evaluating the SDF and gradient grids at the current marker locations using grid sampling with trilinear interpolation. 
We utilize this grid sampling-based SDF feature calculation to achieve a balance between accuracy and computational efficiency.

\section{Comparison to More Related Works}
Here we discuss and compare with more related works on synthesizing human motions in 3D scenes.
\cite{wang_towards_2022, wang_synthesizing_2021} share a similar multi-stage motion synthesis framework of first placing anchor bodies in scenes and then generating in-between trajectories and poses, which requires the pre-specification of the total number of frames. In contrast, our method offers greater flexibility by not constraining the number of frames to generate in advance.
Since \cite{wang_towards_2022} doesn't provide source code, we add the quantitative locomotion comparison with \cite{wang_synthesizing_2021} in their test scenes as shown in Tab.~\ref{tab:compare_longterm}. Our method can generate locomotion results with significantly more natural foot-floor contact and less scene collision.
The generated human motion results from \cite{wang_synthesizing_2021} exhibit artifacts like obvious jittering and foot skating, as evident in the video qualitative comparison in our \href{https://zkf1997.github.io/DIMOS}{project website}.
\begin{table}[ht]
    \centering
    \caption{Quantitative comparison with \cite{wang_synthesizing_2021} on locomotion.}
    \begin{tabular}{lcccccc}
        \hline
       & time $\downarrow$ & avg. dist $\downarrow$ & contact $\uparrow$ & loco pene $\uparrow$\\
        \hline
        Wang etc. \cite{wang_synthesizing_2021}  & 4.00 & \textbf{0.03} & 0.92 & 0.86\\
        Ours & \textbf{3.09} & 0.04 & \textbf{0.99} & \textbf{0.95}\\
        \hline
    \end{tabular}
    \label{tab:compare_longterm}
\end{table}

COUCH \cite{zhang_couch_2022} autoregressively synthesize human motions sitting to chairs satisfying given hand-chair contact constraints. However, COUCH can not handle complex scenes nor generate standing-up motion, and COUCH repeats generating deterministic motion. 
Nevertheless, we quantitatively compare with COUCH on the task of sitting down using their test chairs. The quantitative metrics in Tab.~\ref{tab:compare_couch} show that our method can generate sitting interaction results with faster interaction completion, more natural foot contact, and less human-object penetration. 
We refer to our \href{https://zkf1997.github.io/DIMOS}{project website} for the qualitative comparison of video results.
\begin{table}[ht]
    \centering
    \caption{Quantitative comparison with \cite{zhang_couch_2022} on sitting interactions.}
    \begin{tabular}{lccccc}
        \hline
       & time $\downarrow$ & contact $\uparrow$ & pene. mean $\downarrow$ & pene. max $\downarrow$ \\
        \hline
        COUCH \cite{zhang_couch_2022} & 6.47 & 0.91 & 5.24  & 14.14\\
        Ours & \textbf{3.25} & \textbf{0.97}  & \textbf{1.50} & \textbf{5.89}\\
        \hline
    \end{tabular}
    \label{tab:compare_couch}
\end{table}

%% file: main.bbl
\begin{thebibliography}{10}\itemsep=-1pt

\bibitem{alexanderson_listen_2023}
Simon Alexanderson, Rajmund Nagy, Jonas Beskow, and Gustav~Eje Henter.
\newblock Listen, denoise, action! audio-driven motion synthesis with diffusion
  models.
\newblock {\em ACM Transactions on Graphics (TOG)}, 42(4):1--20, 2023.
\newblock ISBN: 0730-0301 Publisher: ACM New York, NY, USA.

\bibitem{ao_gesturediffuclip_2023}
Tenglong Ao, Zeyi Zhang, and Libin Liu.
\newblock {GestureDiffuCLIP}: {Gesture} diffusion model with {CLIP} latents.
\newblock {\em arXiv preprint arXiv:2303.14613}, 2023.

\bibitem{bergamin_drecon_2019}
Kevin Bergamin, Simon Clavet, Daniel Holden, and James~Richard Forbes.
\newblock {DReCon}: data-driven responsive control of physics-based characters.
\newblock {\em ACM Transactions On Graphics (TOG)}, 38(6):1--11, 2019.
\newblock Publisher: ACM New York, NY, USA.

\bibitem{buttner_machine_2019}
Michael Buttner.
\newblock Machine learning for motion synthesis and character control in games.
\newblock {\em Proc. of I3D 2019}, 2, 2019.

\bibitem{buttner_motion_2015}
Michael Büttner and Simon Clavet.
\newblock Motion matching-the road to next gen animation.
\newblock {\em Proc. of Nucl. ai}, 2015(1):2, 2015.

\bibitem{chang_shapenet_2015}
Angel~X Chang, Thomas Funkhouser, Leonidas Guibas, Pat Hanrahan, Qixing Huang,
  Zimo Li, Silvio Savarese, Manolis Savva, Shuran Song, and Hao Su.
\newblock Shapenet: {An} information-rich 3d model repository.
\newblock {\em arXiv preprint arXiv:1512.03012}, 2015.

\bibitem{dai2017scannet}
Angela Dai, Angel~X. Chang, Manolis Savva, Maciej Halber, Thomas Funkhouser,
  and Matthias Nie{\ss}ner.
\newblock Scannet: Richly-annotated 3d reconstructions of indoor scenes.
\newblock In {\em Proc. Computer Vision and Pattern Recognition (CVPR), IEEE},
  2017.

\bibitem{deschaud2021pariscarla3d}
Jean-Emmanuel Deschaud, David Duque, Jean~Pierre Richa, Santiago
  Velasco-Forero, Beatriz Marcotegui, and François Goulette.
\newblock Paris-carla-3d: A real and synthetic outdoor point cloud dataset for
  challenging tasks in 3d mapping.
\newblock {\em Remote Sensing}, 13(22), 2021.

\bibitem{grabner_what_2011}
Helmut Grabner, Juergen Gall, and Luc Van~Gool.
\newblock What makes a chair a chair?
\newblock pages 1529--1536. IEEE, 2011.

\bibitem{gupta_3d_2011}
Abhinav Gupta, Scott Satkin, Alexei~A Efros, and Martial Hebert.
\newblock From 3d scene geometry to human workspace.
\newblock pages 1961--1968. IEEE, 2011.

\bibitem{habibie_recurrent_2017}
Ikhsanul Habibie, Daniel Holden, Jonathan Schwarz, Joe Yearsley, and Taku
  Komura.
\newblock A recurrent variational autoencoder for human motion synthesis.
\newblock 2017.

\bibitem{hart_formal_1968}
Peter~E Hart, Nils~J Nilsson, and Bertram Raphael.
\newblock A formal basis for the heuristic determination of minimum cost paths.
\newblock {\em IEEE transactions on Systems Science and Cybernetics},
  4(2):100--107, 1968.
\newblock Publisher: IEEE.

\bibitem{hassan_stochastic_2021}
Mohamed Hassan, Duygu Ceylan, Ruben Villegas, Jun Saito, Jimei Yang, Yi Zhou,
  and Michael~J Black.
\newblock Stochastic scene-aware motion prediction.
\newblock pages 11374--11384, 2021.

\bibitem{hassan_resolving_2019}
Mohamed Hassan, Vasileios Choutas, Dimitrios Tzionas, and Michael~J Black.
\newblock Resolving {3D} human pose ambiguities with {3D} scene constraints.
\newblock pages 2282--2292, 2019.

\bibitem{hassan_populating_2021}
Mohamed Hassan, Partha Ghosh, Joachim Tesch, Dimitrios Tzionas, and Michael~J
  Black.
\newblock Populating {3D} scenes by learning human-scene interaction.
\newblock pages 14708--14718, 2021.

\bibitem{hassan_synthesizing_2023}
Mohamed Hassan, Yunrong Guo, Tingwu Wang, Michael Black, Sanja Fidler, and
  Xue~Bin Peng.
\newblock Synthesizing {Physical} {Character}-{Scene} {Interactions}, Feb.
  2023.
\newblock arXiv:2302.00883 [cs].

\bibitem{holden_learned_2020}
Daniel Holden, Oussama Kanoun, Maksym Perepichka, and Tiberiu Popa.
\newblock Learned motion matching.
\newblock {\em ACM Transactions on Graphics (TOG)}, 39(4):53--1, 2020.
\newblock Publisher: ACM New York, NY, USA.

\bibitem{holden_phase-functioned_2017}
Daniel Holden, Taku Komura, and Jun Saito.
\newblock Phase-functioned neural networks for character control.
\newblock {\em ACM Transactions on Graphics (TOG)}, 36(4):1--13, 2017.
\newblock Publisher: ACM New York, NY, USA.

\bibitem{holden_deep_2016}
Daniel Holden, Jun Saito, and Taku Komura.
\newblock A deep learning framework for character motion synthesis and editing.
\newblock {\em ACM Transactions on Graphics (TOG)}, 35(4):1--11, 2016.
\newblock Publisher: ACM New York, NY, USA.

\bibitem{hu_predictive_2020}
Ruizhen Hu, Zihao Yan, Jingwen Zhang, Oliver Van~Kaick, Ariel Shamir, Hao
  Zhang, and Hui Huang.
\newblock Predictive and generative neural networks for object functionality.
\newblock {\em arXiv preprint arXiv:2006.15520}, 2020.

\bibitem{itrch_pynavmesh}
Shekn Itrch.
\newblock pynavmesh: {Python} implementation of path finding algorithm in
  navigation meshes.

\bibitem{juravsky_padl_2022}
Jordan Juravsky, Yunrong Guo, Sanja Fidler, and Xue~Bin Peng.
\newblock {PADL}: {Language}-{Directed} {Physics}-{Based} {Character}
  {Control}.
\newblock In {\em {SIGGRAPH} {Asia} 2022 {Conference} {Papers}}, pages 1--9,
  2022.

\bibitem{kania_trajevae_2021}
Kacper Kania, Marek Kowalski, and Tomasz Trzciński.
\newblock {TrajeVAE}: {Controllable} {Human} {Motion} {Generation} from
  {Trajectories}.
\newblock {\em arXiv preprint arXiv:2104.00351}, 2021.

\bibitem{kleineberg_mesh_to_sdf}
Marian Kleineberg.
\newblock mesh-to-sdf: {Calculate} signed distance fields for arbitrary meshes.

\bibitem{kovar_motion_2008}
Lucas Kovar, Michael Gleicher, and Frédéric Pighin.
\newblock Motion graphs.
\newblock In {\em {ACM} {SIGGRAPH} 2008 classes}, pages 1--10. 2008.

\bibitem{li_ganimator_2022}
Peizhuo Li, Kfir Aberman, Zihan Zhang, Rana Hanocka, and Olga Sorkine-Hornung.
\newblock Ganimator: {Neural} motion synthesis from a single sequence.
\newblock {\em ACM Transactions on Graphics (TOG)}, 41(4):1--12, 2022.
\newblock Publisher: ACM New York, NY, USA.

\bibitem{li_example-based_2023}
Weiyu Li, Xuelin Chen, Peizhuo Li, Olga Sorkine-Hornung, and Baoquan Chen.
\newblock Example-based {Motion} {Synthesis} via {Generative} {Motion}
  {Matching}.
\newblock {\em arXiv preprint arXiv:2306.00378}, 2023.

\bibitem{li_putting_2019}
Xueting Li, Sifei Liu, Kihwan Kim, Xiaolong Wang, Ming-Hsuan Yang, and Jan
  Kautz.
\newblock Putting humans in a scene: {Learning} affordance in 3d indoor
  environments.
\newblock pages 12368--12376, 2019.

\bibitem{ling_character_2020}
Hung~Yu Ling, Fabio Zinno, George Cheng, and Michiel Van De~Panne.
\newblock Character controllers using motion vaes.
\newblock {\em ACM Transactions on Graphics (TOG)}, 39(4):40--1, 2020.
\newblock Publisher: ACM New York, NY, USA.

\bibitem{liu_improving_2015}
Libin Liu, KangKang Yin, and Baining Guo.
\newblock Improving sampling‐based motion control.
\newblock volume~34, pages 415--423. Wiley Online Library, 2015.
\newblock Issue: 2.

\bibitem{liu_sampling-based_2010}
Libin Liu, KangKang Yin, Michiel Van~de Panne, Tianjia Shao, and Weiwei Xu.
\newblock Sampling-based contact-rich motion control.
\newblock In {\em {ACM} {SIGGRAPH} 2010 papers}, pages 1--10. 2010.

\bibitem{mahmood_amass_2019}
Naureen Mahmood, Nima Ghorbani, Nikolaus~F Troje, Gerard Pons-Moll, and
  Michael~J Black.
\newblock {AMASS}: {Archive} of motion capture as surface shapes.
\newblock pages 5442--5451, 2019.

\bibitem{merel_neural_2018}
Josh Merel, Leonard Hasenclever, Alexandre Galashov, Arun Ahuja, Vu Pham, Greg
  Wayne, Yee~Whye Teh, and Nicolas Heess.
\newblock Neural probabilistic motor primitives for humanoid control.
\newblock {\em arXiv preprint arXiv:1811.11711}, 2018.

\bibitem{pavlakos_expressive_2019}
Georgios Pavlakos, Vasileios Choutas, Nima Ghorbani, Timo Bolkart, Ahmed~AA
  Osman, Dimitrios Tzionas, and Michael~J Black.
\newblock Expressive body capture: 3d hands, face, and body from a single
  image.
\newblock pages 10975--10985, 2019.

\bibitem{peng_deepmimic_2018}
Xue~Bin Peng, Pieter Abbeel, Sergey Levine, and Michiel Van~de Panne.
\newblock Deepmimic: {Example}-guided deep reinforcement learning of
  physics-based character skills.
\newblock {\em ACM Transactions On Graphics (TOG)}, 37(4):1--14, 2018.
\newblock Publisher: ACM New York, NY, USA.

\bibitem{peng_ase_2022}
Xue~Bin Peng, Yunrong Guo, Lina Halper, Sergey Levine, and Sanja Fidler.
\newblock Ase: {Large}-scale reusable adversarial skill embeddings for
  physically simulated characters.
\newblock {\em ACM Transactions On Graphics (TOG)}, 41(4):1--17, 2022.
\newblock Publisher: ACM New York, NY, USA.

\bibitem{peng_sfv_2018}
Xue~Bin Peng, Angjoo Kanazawa, Jitendra Malik, Pieter Abbeel, and Sergey
  Levine.
\newblock Sfv: {Reinforcement} learning of physical skills from videos.
\newblock {\em ACM Transactions On Graphics (TOG)}, 37(6):1--14, 2018.
\newblock Publisher: ACM New York, NY, USA.

\bibitem{petrovich_action-conditioned_2021}
Mathis Petrovich, Michael~J Black, and Gül Varol.
\newblock Action-conditioned {3D} human motion synthesis with transformer
  {VAE}.
\newblock pages 10985--10995, 2021.

\bibitem{petrovich_temos_2022}
Mathis Petrovich, Michael~J Black, and Gül Varol.
\newblock {TEMOS}: {Generating} diverse human motions from textual
  descriptions.
\newblock pages 480--497. Springer, 2022.

\bibitem{punnakkal_babel_2021}
Abhinanda~R Punnakkal, Arjun Chandrasekaran, Nikos Athanasiou, Alejandra
  Quiros-Ramirez, and Michael~J Black.
\newblock {BABEL}: bodies, action and behavior with {English} labels.
\newblock pages 722--731, 2021.

\bibitem{savva_scenegrok_2014}
Manolis Savva, Angel~X Chang, Pat Hanrahan, Matthew Fisher, and Matthias
  Nießner.
\newblock {SceneGrok}: {Inferring} action maps in {3D} environments.
\newblock {\em ACM transactions on graphics (TOG)}, 33(6):1--10, 2014.
\newblock Publisher: ACM New York, NY, USA.

\bibitem{schulman_proximal_2017}
John Schulman, Filip Wolski, Prafulla Dhariwal, Alec Radford, and Oleg Klimov.
\newblock Proximal policy optimization algorithms.
\newblock {\em arXiv preprint arXiv:1707.06347}, 2017.

\bibitem{snook_simplified_2000}
Greg Snook.
\newblock Simplified {3D} movement and pathfinding using navigation meshes.
\newblock {\em Game programming gems}, 1(1):288--304, 2000.
\newblock Publisher: Charles River Media Newton Centre, MA, USA.

\bibitem{starke_deepphase_2022}
Sebastian Starke, Ian Mason, and Taku Komura.
\newblock Deepphase: {Periodic} autoencoders for learning motion phase
  manifolds.
\newblock {\em ACM Transactions on Graphics (TOG)}, 41(4):1--13, 2022.
\newblock Publisher: ACM New York, NY, USA.

\bibitem{starke_neural_2019}
Sebastian Starke, He Zhang, Taku Komura, and Jun Saito.
\newblock Neural state machine for character-scene interactions.
\newblock {\em ACM Trans. Graph.}, 38(6):209--1, 2019.

\bibitem{starke_local_2020}
Sebastian Starke, Yiwei Zhao, Taku Komura, and Kazi Zaman.
\newblock Local motion phases for learning multi-contact character movements.
\newblock {\em ACM Transactions on Graphics (TOG)}, 39(4):54--1, 2020.
\newblock Publisher: ACM New York, NY, USA.

\bibitem{straub_replica_2019}
Julian Straub, Thomas Whelan, Lingni Ma, Yufan Chen, Erik Wijmans, Simon Green,
  Jakob~J Engel, Raul Mur-Artal, Carl Ren, and Shobhit Verma.
\newblock The {Replica} dataset: {A} digital replica of indoor spaces.
\newblock {\em arXiv preprint arXiv:1906.05797}, 2019.

\bibitem{sutton_introduction_1998}
Richard~S Sutton and Andrew~G Barto.
\newblock {\em Introduction to reinforcement learning}, volume 135.
\newblock MIT press Cambridge, 1998.

\bibitem{tang_real-time_2022}
Xiangjun Tang, He Wang, Bo Hu, Xu Gong, Ruifan Yi, Qilong Kou, and Xiaogang
  Jin.
\newblock Real-time controllable motion transition for characters.
\newblock {\em ACM Transactions on Graphics (TOG)}, 41(4):1--10, 2022.
\newblock Publisher: ACM New York, NY, USA.

\bibitem{tessler_calm_2023}
Chen Tessler, Yoni Kasten, Yunrong Guo, Shie Mannor, Gal Chechik, and Xue~Bin
  Peng.
\newblock {CALM}: {Conditional} {Adversarial} {Latent} {Models} for
  {Directable} {Virtual} {Characters}.
\newblock In {\em {ACM} {SIGGRAPH} 2023 {Conference} {Proceedings}}, pages
  1--9, 2023.

\bibitem{tevet_motionclip_2022}
Guy Tevet, Brian Gordon, Amir Hertz, Amit~H Bermano, and Daniel Cohen-Or.
\newblock Motionclip: {Exposing} human motion generation to clip space.
\newblock pages 358--374. Springer, 2022.

\bibitem{tevet_human_2022}
Guy Tevet, Sigal Raab, Brian Gordon, Yonatan Shafir, Daniel Cohen-Or, and
  Amit~H Bermano.
\newblock Human motion diffusion model.
\newblock {\em arXiv preprint arXiv:2209.14916}, 2022.

\bibitem{tseng_edge_2022}
Jonathan Tseng, Rodrigo Castellon, and C~Karen Liu.
\newblock {EDGE}: {Editable} {Dance} {Generation} {From} {Music}.
\newblock {\em arXiv preprint arXiv:2211.10658}, 2022.

\bibitem{wang_towards_2022}
Jingbo Wang, Yu Rong, Jingyuan Liu, Sijie Yan, Dahua Lin, and Bo Dai.
\newblock Towards diverse and natural scene-aware 3d human motion synthesis.
\newblock pages 20460--20469, 2022.

\bibitem{wang_synthesizing_2021}
Jiashun Wang, Huazhe Xu, Jingwei Xu, Sifei Liu, and Xiaolong Wang.
\newblock Synthesizing long-term 3d human motion and interaction in 3d scenes.
\newblock pages 9401--9411, 2021.

\bibitem{wang_combining_2019}
Zhiyong Wang, Jinxiang Chai, and Shihong Xia.
\newblock Combining recurrent neural networks and adversarial training for
  human motion synthesis and control.
\newblock {\em IEEE transactions on visualization and computer graphics},
  27(1):14--28, 2019.
\newblock Publisher: IEEE.

\bibitem{yuan_physdiff_2022}
Ye Yuan, Jiaming Song, Umar Iqbal, Arash Vahdat, and Jan Kautz.
\newblock {PhysDiff}: {Physics}-{Guided} {Human} {Motion} {Diffusion} {Model}.
\newblock {\em arXiv preprint arXiv:2212.02500}, 2022.

\bibitem{zhang_learning_2023}
Haotian Zhang, Ye Yuan, Viktor Makoviychuk, Yunrong Guo, Sanja Fidler, Xue~Bin
  Peng, and Kayvon Fatahalian.
\newblock Learning {Physically} {Simulated} {Tennis} {Skills} from {Broadcast}
  {Videos}.
\newblock {\em ACM Transactions on Graphics (TOG)}, 42(4):1--14, 2023.
\newblock ISBN: 0730-0301 Publisher: ACM New York, NY, USA.

\bibitem{Zhang:ICCV:2021}
Siwei Zhang, Yan Zhang, Federica Bogo, Pollefeys Marc, and Siyu Tang.
\newblock Learning motion priors for 4d human body capture in 3d scenes.
\newblock In {\em International Conference on Computer Vision (ICCV)}, Oct.
  2021.

\bibitem{zhang_place_2020}
Siwei Zhang, Yan Zhang, Qianli Ma, Michael~J Black, and Siyu Tang.
\newblock {PLACE}: {Proximity} learning of articulation and contact in {3D}
  environments.
\newblock pages 642--651. IEEE, 2020.

\bibitem{zhang_couch_2022}
Xiaohan Zhang, Bharat~Lal Bhatnagar, Sebastian Starke, Vladimir Guzov, and
  Gerard Pons-Moll.
\newblock Couch: towards controllable human-chair interactions.
\newblock pages 518--535. Springer, 2022.

\bibitem{zhang_we_2021}
Yan Zhang, Michael~J Black, and Siyu Tang.
\newblock We are more than our joints: {Predicting} how 3d bodies move.
\newblock pages 3372--3382, 2021.

\bibitem{zhang_generating_2020}
Yan Zhang, Mohamed Hassan, Heiko Neumann, Michael~J Black, and Siyu Tang.
\newblock Generating 3d people in scenes without people.
\newblock pages 6194--6204, 2020.

\bibitem{zhang_wanderings_2022}
Yan Zhang and Siyu Tang.
\newblock The wanderings of odysseus in {3D} scenes.
\newblock pages 20481--20491, 2022.

\bibitem{zhao_compositional_2022}
Kaifeng Zhao, Shaofei Wang, Yan Zhang, Thabo Beeler, and Siyu Tang.
\newblock Compositional human-scene interaction synthesis with semantic
  control.
\newblock pages 311--327. Springer, 2022.

\bibitem{zheng_gimo_2022}
Yang Zheng, Yanchao Yang, Kaichun Mo, Jiaman Li, Tao Yu, Yebin Liu, C~Karen
  Liu, and Leonidas~J Guibas.
\newblock Gimo: {Gaze}-informed human motion prediction in context.
\newblock pages 676--694. Springer, 2022.

\bibitem{zinno_ml_2019}
Fabio Zinno.
\newblock Ml tutorial day: {From} motion matching to motion synthesis, and all
  the hurdles in between.
\newblock {\em Proc. of GDC 2019}, 2, 2019.

\end{thebibliography}
